\documentclass{article} %
\usepackage{iclr2027_conference,times}

\usepackage{amsmath,amssymb,amsfonts}
\usepackage{graphicx}
\usepackage{url}
\usepackage{booktabs}
\usepackage{tabularx}
\usepackage{array}
\usepackage{chngcntr}
\usepackage{placeins}
\usepackage{hyperref}

\hypersetup{
    hidelinks,
    pdftitle={Where Predictive Supervision Goes Shapes What VLA Policies Learn},
    pdfauthor={Hanseul Kim, Jewon Yeom, Youngjoon Jeong, Minsoo Jo, Taesup Kim}
}

\newcolumntype{Y}{>{\raggedright\arraybackslash}X}

\title{Where Predictive Supervision Goes\\Shapes What VLA Policies Learn}

\author{Hanseul Kim, Jewon Yeom, Youngjoon Jeong, Minsoo Jo, Taesup Kim\thanks{Corresponding author.} \\
Graduate School of Data Science\\
Seoul National University\\
\texttt{\{k1seul, jewon0908, af1014, gsds.minsooj, taesup.kim\}@snu.ac.kr} \\
}

\iclrfinalcopy %
\begin{document}

\maketitle
\lhead{Preprint. Under review.}

\addtocontents{toc}{\protect\setcounter{tocdepth}{-1}}

\begin{abstract}
Future prediction is increasingly used to improve vision-language-action (VLA)
policies, based on the premise that anticipating scene evolution encourages
representations useful for control. However, forecast quality alone does not
establish that a policy has learned a better representation for action. This
distinction matters under distribution shift, where successful control depends
on preserving spatial state and likely scene change beyond familiar
configurations. We study what determines whether predictive supervision
improves the visual representation used by a VLA policy. Through controlled
comparisons with matched target constructions, prediction horizons, and
training conditions, we find that different prediction interfaces produce
markedly different forecasts and visual representations, including in the
spatial, dynamics, and action information that transfers beyond familiar
scenes. We trace these differences to how predictive errors shape the policy's
visual stream. Consistent with this controlled finding, VLA policies trained
with more direct, scene-matched future supervision show stronger robustness
under simulated and physical distribution shifts. Together, our results frame
future prediction as a representation-learning design problem whose value for
control depends on whether its supervision reaches the representations through
which the policy acts.
\end{abstract}

\section{Introduction}

Vision-language-action (VLA) models have become capable manipulation policies
across diverse tasks, datasets, and embodiments~\citep{Brohan2023-fh,
Kim2024-xd, Intelligence2025-jz}. Yet policies that perform well
on standard tasks remain brittle when familiar objects move or the surrounding
geometry changes. Controlled OOD benchmarks expose large gaps under changes in
object identity, spatial configuration, camera viewpoint, language, and robot
state~\citep{Zhou2026-gm, Fei2025-jb, Morgan2026-zf}. Such changes
are unavoidable in deployment. Robust action therefore requires visual
representations that preserve actionable spatial structure and how the scene
may change, rather than only recognizing what is currently visible.

Future prediction offers a natural source of such supervision. Robot
trajectories already contain observations of how scenes evolve through
interaction, allowing policies to learn from future frames, latent states,
visual features, or motion alongside the action objective~\citep{Wu2023-ps,
Zhao2025-br, Zhang2025-yd, Li2026-fomovla, Syed2026-ahead}. The appeal is
straightforward: anticipating the consequences of interaction may encourage
the policy to represent the state and dynamics needed for control, including
those that must transfer beyond familiar configurations.

Yet adding a future-prediction objective does not ensure that a policy learns
a better representation for control. Predictive accuracy need not align with
downstream control performance~\citep{Lambert2020-objective}, and representation
objectives can preserve different task-relevant factors~\citep{Zhang2021-invariant}.
In a VLA, one possible source of this mismatch is architectural. The auxiliary
objective may be optimized through a pathway that only weakly shapes the visual
stream used for action. Prediction quality alone therefore does not reveal whether
the policy has learned the spatial and temporal structure needed to act when the
scene changes. This leaves a
fundamental question unresolved: what determines whether future prediction
actually improves the representation through which a VLA controls the robot?

We study future prediction as a representation-learning problem. Our central
finding is that a common future-prediction objective can teach substantially different
visual representations depending on how its errors reach the policy's visual
stream. Across controlled representation learning and VLA training, the most
transferable features emerge when scene-matched predictive supervision
directly shapes the spatial visual stream used for action. The resulting
differences remain consequential under simulated and physical distribution
shifts. Predictive supervision is therefore not an auxiliary capability whose
value can be judged from its forecast alone. Where it enters the policy shapes
what the policy learns and how robustly that knowledge supports control.

Our contributions are:
\begin{itemize}
    \item We separate forecast output from policy representation, showing that
    matched future objectives can produce substantially different visual
    representations.
    \item We establish that where predictive supervision enters the policy
    determines whether action-relevant spatial and temporal information remains
    accessible in its visual representation.
    \item We show that this representation-level distinction remains
    consequential for VLA robustness under simulated and physical distribution
    shifts.
\end{itemize}

\section{Related Work}

\subsection{Predictive Supervision in Robot Policies}

Future-predictive robot policies differ in both the target they predict and the
policy state directly optimized by that objective. Targets include future
frames, image codes, latent states, visual features, and motion
representations~\citep{Wu2023-ps, Cheang2024-cc, Wang2025-ta, Zhu2025-fq,
Lin2025-hifvla}. Predictions may appear as intermediate outputs, learned
carriers, or states supplied by a separate world model. This distinction
determines whether forecast error directly trains the policy representation
used for control. Appendix~\ref{app:prior-interfaces} compares representative interfaces by this
connection.

CoT-VLA generates future image codes at autoregressive output
positions~\citep{Zhao2025-br}. FLARE, DreamVLA, World Guidance, and HiF-VLA
attach latent or motion targets to learned carriers that interact with visual
tokens through attention~\citep{Zheng2025-du, Zhang2025-yd,
Su2026-worldguidance, Lin2025-hifvla}. VLA-JEPA uses latent state prediction
during video pretraining, while FutureVLA aligns downstream VLA states with
pretrained joint visuomotor embeddings~\citep{Sun2026-fb,
Xu2026-futurevla}. These approaches transfer predictive structure into control
without attaching each forecast loss directly to the corresponding policy
patch.

FoMoVLA jointly supervises future features and sparse point
trajectories~\citep{Li2026-fomovla}. AHEAD instead forecasts patch-aligned features
in a separate world model around a frozen VLA~\citep{Syed2026-ahead}. Prior
methods therefore establish the value of future prediction and spatial
supervision, but vary targets, modules, or training stages together. They do
not isolate whether a matched future-prediction objective teaches a different
policy representation when only its route into the visual stream changes. We
hold the target construction, prediction horizon, readout count, backbone, and
optimization fixed while varying spatial
address and direct coupling.

\subsection{Supervision Placement in Visual Representation Learning}

Representation learning provides a complementary view by asking which hidden
states an objective directly trains. Learned class and query tokens offer
flexible readouts without fixed patch correspondence~\citep{Dosovitskiy2020-hz,
Carion2020-hd, Li2023-tp}. Attention and positional structure can give these
tokens a spatial address, but address alone does not specify how supervision
reaches patch representations. Head placement can also change where
transferable features emerge~\citep{Chen2020-simclr, Ren2025-deepmim,
Alkin2025-mimrefiner}.

Spatial and temporal objectives provide complementary evidence. iBOT applies
masked self-distillation at patch tokens, and GLaD aligns VLA visual positions
with a geometry-aware teacher~\citep{Zhou2021-ey, Guo2025-glad}. Dense
Predictive Coding, SPR, DINO-WM, DINO-Foresight, and V-JEPA 2 instead learn
from future representations~\citep{Han2019-dpc, Schwarzer2021-spr,
Zhou2025-dinowm, Karypidis2025-dinoforesight, Assran2025-vjepa2}. Our question
lies at their intersection: whether attaching the same temporal target to
different positions in a trainable policy visual stream changes what that
stream learns.

\section{Problem Formulation}

\subsection{Future Prediction as Auxiliary Supervision}

Let $o_t$ and $o_{t+h}$ be observations $h$ steps apart. The online
visual encoder produces $V_t=[v_1,\ldots,v_N]$, and its EMA target encoder
produces $Y_\tau=[y_{\tau,1},\ldots,y_{\tau,N}]$ for
$\tau\in\{t,t+h\}$. The patchwise objective is
\begin{equation}
\begin{aligned}
V_t &= E_{\theta_v}(o_t), \qquad
Y_t = E_{\bar{\theta}_v}(o_t), \qquad
Y_{t+h} = E_{\bar{\theta}_v}(o_{t+h}),\\
r_i &= \operatorname{sg}\!\left(y_{t+h,i}-y_{t,i}\right), \qquad
\mathcal L_{\mathrm{fut}}
= \frac{1}{N}\sum_{i=1}^{N}
\mathcal D\!\left(H_\psi(z_i),r_i\right).
\end{aligned}
\end{equation}
Here, $\operatorname{sg}$ stops gradients, $r_i$ is the residual target at
patch $i$, $z_i$ is the interface-selected current token, $H_\psi$ is the
token-wise head, $\mathcal D$ is the per-target loss, and $N$ is the number of
visual patches. RQ1 and RQ2 optimize $\mathcal L_{\mathrm{fut}}$ alone, while
RQ3 uses $\mathcal L_{\mathrm{act}}+\lambda\mathcal L_{\mathrm{fut}}$, where
$\lambda$ weights future prediction. The equation gives the patchwise form.
The native $M=16$ special-token VLA interface instead uses region-pooled
targets (Appendix~\ref{app:vla-policy-training-evaluation}). Comparisons fix
the target, horizon, backbone, and optimization to isolate the interface.

\subsection{Prediction Interfaces as Credit Routes}

Let $S_t=[s_1,\ldots,s_M]$ denote $M$ special tokens and a tilde denote a token
after contextual mixing. The special-token interface reads $z_i=\tilde s_i$,
while the vision-token interface reads $z_i=\tilde v_i$ at the corresponding
patch. Forecast credit denotes the gradient induced on visual tokens by an
individual forecast error. A special-token error reaches patches only through
attention-mediated mixing. A vision-token error also passes through mixing,
but its residual stream contains an explicit same-position identity term
(Appendix~\ref{app:derivation-gradients}).

\begin{figure*}[t]
    \centering
    \includegraphics[width=\textwidth,trim=17 9 13 12,clip]{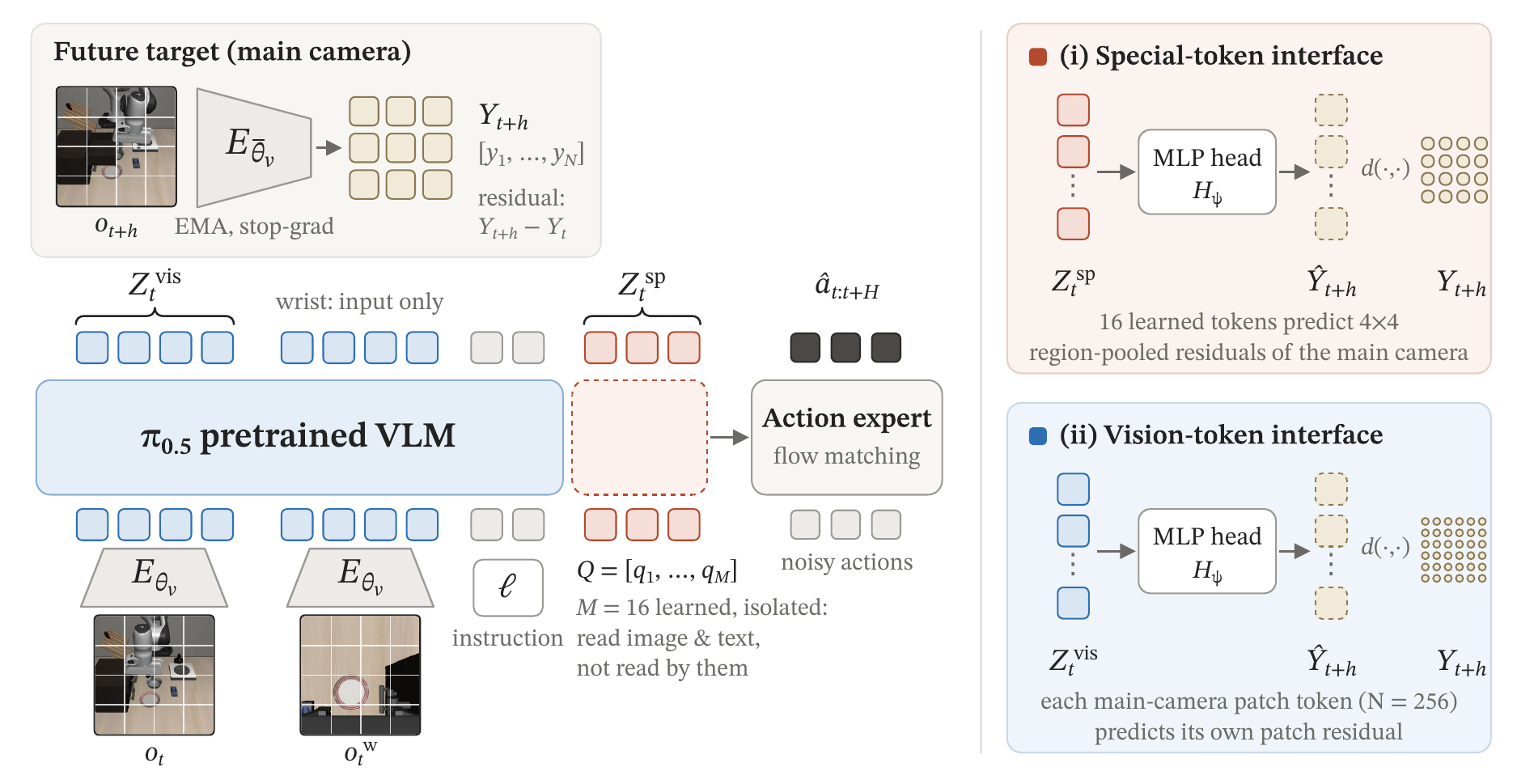}
    \caption{
    \textbf{Downstream VLA realization of future-feature prediction used in RQ3.}
    A momentum encoder supplies main-camera targets while the pretrained VLM
    and action expert are shared. The special-token interface predicts from
    special tokens, whereas the vision-token interface predicts at
    main-camera patch positions and retains a same-position credit route.
    }
    \label{fig:vla-integration}
\end{figure*}

Special tokens may acquire spatial correspondence through attention or
positional structure, but their readout does not guarantee it. The distinction
therefore concerns how spatially matched forecast credit reaches the visual
stream, not whether either token type can contain visual information.
Figure~\ref{fig:vla-integration} illustrates the two routes in the downstream
policy.

The formulation above separates what future state is predicted from where the
corresponding supervision is attached within the policy. Holding the target
construction and prediction horizon fixed allows the prediction interface to
be examined as an independent factor in policy learning, providing the basis
for the questions below.

\subsection{Research Questions}
\label{sec:research-questions}

\paragraph{RQ1. Does the prediction interface shape what the visual stream learns?}
Prediction interfaces are often treated as implementation choices. If they
change the information available to the policy, however, they become central
to understanding how future prediction shapes the representation used for
control.

\paragraph{RQ2. Why do prediction interfaces lead to different representations?}
Showing that two interfaces learn differently is not enough to guide model
design. Identifying the source of that difference is necessary for the finding
to generalize beyond the particular architectures being compared.

\paragraph{RQ3. Do these differences improve VLA robustness?}
Representational differences matter only if they influence policy behavior.
Testing whether their effects persist under distribution shift connects the
controlled analysis to the broader challenge of robust generalization in VLA
policies.

\section{The Prediction Interface Shapes What the Visual Stream Learns}
\label{sec:future-prediction-representations}

We first test this question in the prediction-only setting shown in Appendix
Figure~\ref{fig:controlled-forecasting-interfaces}. We compare special-token
and vision-token interfaces under a matched learning problem. The architecture
and training details appear in Appendix~\ref{app:controlled-forecasting-setting},
with the forecast and probe protocols in
Appendices~\ref{app:decoded-forecast-evaluation}
and~\ref{app:frozen-feature-probes}.

\begin{figure*}[t]
    \centering
    \includegraphics[width=0.9\textwidth]{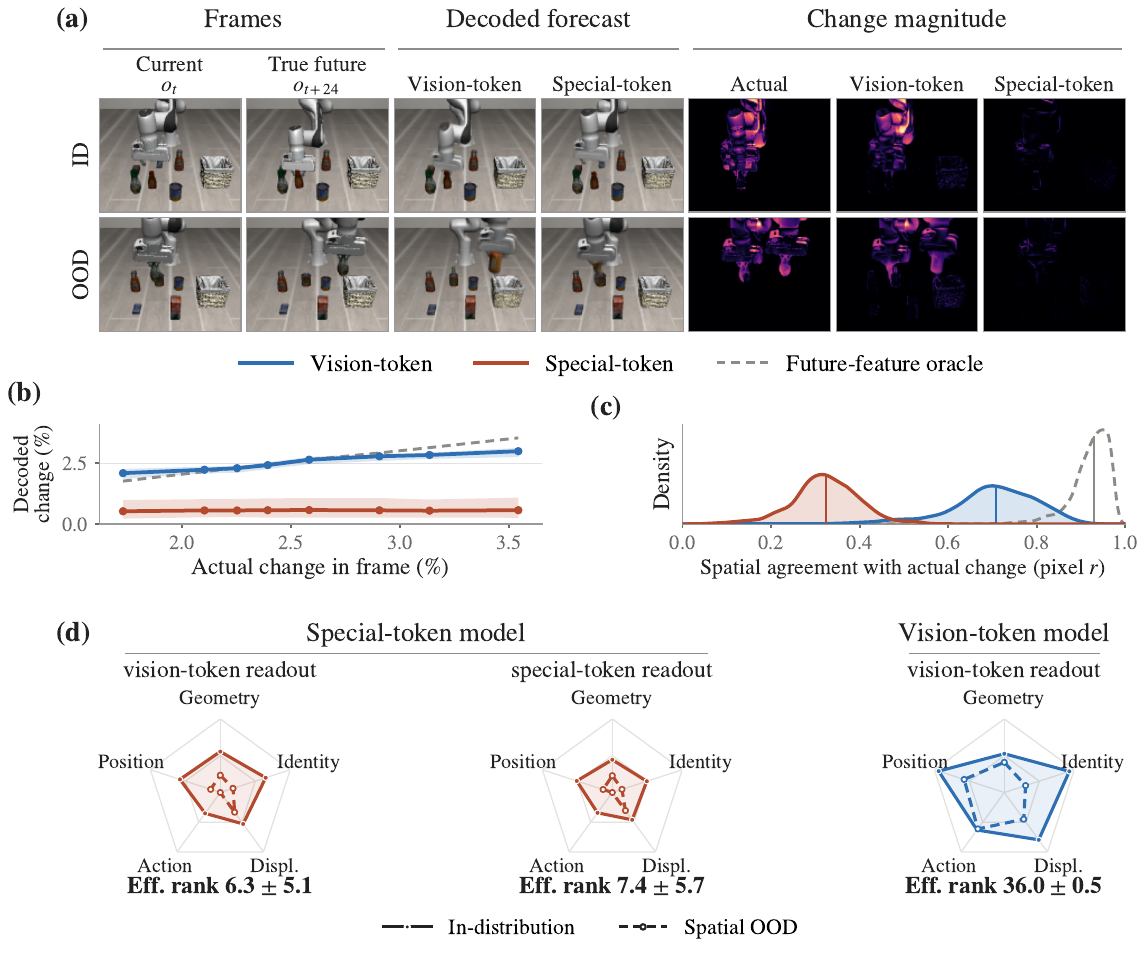}
    \caption{
    \textbf{Vision-token forecasting captures scene-specific future change and
    preserves transferable visual features.}
    \textbf{(a)} Near-center and spatial-tail examples show decoded forecasts
    and change maps.
    \textbf{(b)} Decoded and actual scene-change magnitudes.
    \textbf{(c)} Their per-example spatial correlation.
    \textbf{(d)} Frozen probes of state, dynamics, actions, and geometry.
    Solid and dashed outlines mark near-center and spatial-tail results, split
    at each task's median distance from its mean target position. Effective
    rank appears below each radar.
    }
    \label{fig:rq1-forecast-probes}
\end{figure*}

\paragraph{Vision-token forecasts represent scene-specific future change.}
We examine what future change each interface actually predicts.
Figure~\ref{fig:rq1-forecast-probes}(a--c) evaluates forecast content
using a frozen pixel decoder trained only on real-frame features from disjoint
episodes. The decoded current and true future features define a copy baseline
and a future-feature oracle, and forecast quality measures how much of the gap
between them is closed by the prediction.

The decoded forecasts separate the two interfaces clearly. Vision-token
predictions track both the magnitude and spatial location of the actual scene
change, while special-token predictions remain close to a copy of the current
observation. Quantitatively, vision-token forecasts close $45.0\%$ of the
full-image gap and $56.4\%$ on moving pixels, whereas special-token forecasts
close almost none. The same qualitative ordering holds for both near-center and spatial-tail
samples.

\paragraph{Vision-token forecasting preserves transferable state and dynamics.}
Figure~\ref{fig:rq1-forecast-probes}(d) asks whether the difference in
forecast content is also reflected in the learned representation. We freeze
each encoder and train ridge probes to read current object position and
identity, future displacement, upcoming actions, and object-centered geometry.
The probes are fitted on clips from the inner half of each task's
target-position distribution, defined by distance to the task-specific mean,
and evaluated on both this near-center set and the outer spatial-tail set from
held-out episodes. This tests probe-level spatial transfer rather than an
unseen distribution for the encoder. For the special-token model, we probe the
visual and special-token streams separately to test whether the special
tokens retain information missing from the visual stream.

On the spatial-tail evaluation set, vision-token features continue to support
linear decoding of current state, future dynamics, and upcoming action, even
though the probes are fitted only on near-center clips. The contrast is
strongest for upcoming action, for which vision-token features retain a linear
readout of $R^2=0.612$ while the special-token visual stream does not support a
reliable readout. Object position, future displacement, and object-centered
geometry show the same ordering, indicating that the interface affects a broad
set of spatial and action-relevant properties rather than a single probe.
Identity follows the same ordering but degrades for every model, suggesting
that the interface advantage is strongest for spatial and action-related
structure rather than uniform across representation properties.

Linear probes on the special-token stream do not recover the state and
dynamics that are weakly represented in the visual stream. Their performance
remains poor on the spatial-tail evaluation set, indicating that these
properties are not linearly accessible from the special tokens either. The
same contrast appears in the effective rank of the visual-token covariance.
The vision-token representation is broader and more stable, while both
readout sites in the special-token model remain lower rank. The prediction
interface therefore changes what the visual stream learns, not merely where
the forecast is decoded.

\paragraph{Matched future targets can produce different visual representations across prediction interfaces (RQ1).}
Future prediction is therefore not an interface-agnostic objective. Attaching
the forecast to vision tokens keeps scene change, spatial state, and future
action accessible in the stream used for control. Moving the same prediction
target to special tokens can leave that stream weak. Linear probes on the
special-token stream do not recover the missing information. What the policy
learns depends not only on what future is predicted, but also
on where predictive supervision enters the representation. These results are
not simply a failure of the momentum target.
Appendix~\ref{app:frozen-teacher-control} shows that special tokens can
forecast well with a frozen pretrained target or stable spatial indexing, but
better forecasting does not by itself close the gap in transferable visual
features.

\section{Direct Forecast Coupling Supports Transferable Visual Features}
\label{sec:spatial-credit-assignment}

\begin{figure*}[t]
    \centering
    \includegraphics[width=0.9\linewidth]{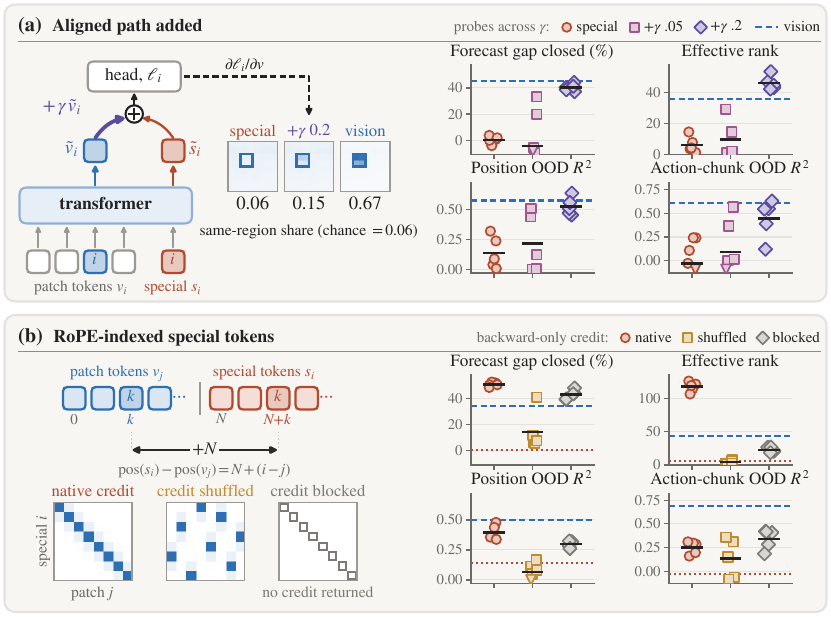}
    \caption{
    \textbf{Direct aligned coupling recovers transferable visual features.}
    \textbf{(a)} Aligned-path construction, same-region gradient share at
    initialization, and probe outcomes across path strengths $\gamma$.
    \textbf{(b)} RoPE-indexed routing and probe outcomes when
    special-token-to-patch gradients are preserved, shuffled, or blocked under
    identical forward computation. Dots are seeds and black bars are means. Dashed blue and
    dotted red lines denote the vision-token and unindexed special-token
    references.
    }
    \label{fig:rq2-credit}
\end{figure*}

RQ1 establishes a representation gap, but it leaves open whether
special-token forecasting merely learns a weaker predictor or trains the
visual stream through a less effective route. We therefore examine how each
spatial forecast loss trains the visual stream. Figure~\ref{fig:rq2-credit}
separates whether the loss has a direct path to the corresponding patch from
whether its gradient reaches the matching image region.

Panel (a) adds a direct aligned path. Panel (b) instead preserves, shuffles, or
blocks attention-mediated gradients without changing the forward computation.
We quantify spatial alignment using the same-region gradient share, the
fraction of the visual-token gradient norm that falls within the corresponding
image region. Full definitions are provided in
Appendix~\ref{app:gradient-delivery}.

\paragraph{The vision-token interface directly couples each forecast error to its patch.}
At initialization, the gradient from an unindexed special-token forecast is no
more concentrated in the corresponding image region than chance. Vision-token
forecasting instead includes a same-position residual connection, so every
forecast error has a direct gradient path to its matching patch. This
difference exists before either model has learned a routing pattern, motivating
the aligned-path intervention in Figure~\ref{fig:rq2-credit}(a).

\paragraph{Adding an aligned path rescues the special-token representation.}
We add the corresponding visual token to each special token,
\[
z_i^{(\gamma)}=\tilde s_i+p_i+\gamma\tilde v_i,
\qquad
\hat r_i=H_\psi\!\left(z_i^{(\gamma)}\right).
\]
Here, $p_i$ is the learned position query for special token $i$, and $\gamma$
controls the strength of the aligned visual path.
The added term provides the direct same-position coupling absent from the
original special-token interface. A very weak path produces seed-dependent
outcomes, but a moderate path consistently prevents the low-rank, near-copy solution
and recovers the forecast, representation breadth, and spatial transfer toward
the vision-token model (Figure~\ref{fig:rq2-credit}(a)). Even this partial
aligned coupling can move the model out of this solution, while
native vision-token readout retains advantages on several measures. The
intervention changes both what the prediction head reads and how forecast
gradients reach the visual stream, so this rescue alone does not separate their
contributions. Removing the path at evaluation preserves part of the forecast
recovery, showing that training leaves a persistent representation change
rather than relying entirely on the path at readout time
(Appendix~\ref{app:aligned-path-rescue}).

\paragraph{Changing only the backward route isolates learning through special-token attention.}
To isolate gradient delivery, we construct a diagnostic $M=N$ special-token
model with a stable one-to-one
attention correspondence between special token $i$ and patch $i$. Rotary
positional embeddings (RoPE) produce this correspondence in every seed,
whereas the unindexed RQ1 model has no stable special-token-to-patch route
(Figure~\ref{fig:rq2-credit}(b)). We also give the special-token and visual
streams separate Transformer weights, ensuring that the visual stream can be
trained only by gradients returned through special-token attention.

The forward computation is identical across all three conditions. The native
condition returns the gradient through the patches attended by the special
tokens. The blocked condition stops this gradient before it reaches the visual
stream, providing an untrained visual-stream reference. The shuffled condition
redirects the same gradient to mismatched patches, testing whether its spatial
destination matters. Any difference among these conditions therefore arises
from the backward route rather than from the information available during
prediction.

\paragraph{Correctly routed attention gradients train the visual stream but do not recover action transfer.}
Comparing the native and blocked conditions first establishes that the returned
gradient is an effective learning signal. Allowing it to reach the visual
stream improves the decoded forecast, effective rank, and position decoding.
It does not, however, improve action decoding on the spatial-tail evaluation
set. The same limitation appears relative to the vision-token reference. The
native condition produces a better forecast and a higher feature rank, yet
retains substantially weaker action transfer. Spatial address still matters,
as redirecting the gradient to mismatched patches collapses both forecast and
visual readouts toward the unindexed special-token model
(Figure~\ref{fig:rq2-credit}(b)). In this untied diagnostic,
attention-returned gradients can therefore shape the visual representation,
but correct addressing alone does not reproduce the transferable,
action-readable features produced by direct vision-token coupling. When the
two streams share Transformer weights, route-enabled and blocked conditions
remain comparable in action transfer, indicating that shared parameter
updates provide an additional learning route
(Appendix~\ref{app:rope-special-credit}).

\paragraph{How predictive supervision reaches the visual stream shapes what becomes transferable (RQ2).}
In the controlled setting, direct same-position coupling provides the most
reliable route from future prediction to transferable visual features.
Correctly addressed attention-returned gradients shape the visual stream, but
do not by themselves reproduce the action transfer obtained through direct
coupling. Shared parameters provide an additional learning route, so the
broader conclusion is that the prediction interface determines which
representations future supervision trains.

\section{Forecast Routing and Target Content Shape VLA Robustness}
\label{sec:vla-evaluation}

RQ1 and RQ2 show that the route of predictive supervision changes what
the visual stream learns. The remaining question is whether this distinction
affects the robustness of a policy trained jointly for prediction and action.
If predictive supervision supports control through the visual representation,
its benefit should depend both on how supervision reaches that representation
and on whether it carries scene-matched future information. We examine this
connection under simulated and physical distribution shifts.

\paragraph{Policy variants.}
All variants fine-tune the same pretrained $\pi_{0.5}$ checkpoint on matched
LIBERO demonstrations. Future-feature prediction is integrated into the
main-camera stream while the wrist input and action expert remain unchanged
(Figure~\ref{fig:vla-integration}). The central comparison places
the native $M=16$ special-token and vision-token interfaces against an
action-only baseline. Two $M=N=256$ special-token controls then separate
token count and patchwise target granularity from stable spatial addressing.
Both predict one target per patch, while only the anchored-index variant fixes
the special-token-to-patch offsets, testing whether stable spatial addressing
recovers part of the gap. Finally, a shuffled-future vision-token
control preserves the direct route while removing scene-matched predictive
content, testing whether that route is useful only when it carries relevant
future change. Appendix~\ref{app:vla-policy-training-evaluation} describes the variants, and
Appendix~\ref{app:vla-m256-delivery} reports their routing measurements.

\paragraph{Evaluation protocols.}
We evaluate robustness in complementary simulated and physical settings.
Simulation provides broad, controlled coverage across four LIBERO suites under
standard evaluation and five LIBERO-PRO perturbation families, with three
training seeds per variant~\citep{Liu2023-uj,Zhou2026-gm}. We report
object-position shift separately because it most directly tests the
spatial-addressing hypothesis, alongside mean performance and retention across
all perturbations. Appendix~\ref{app:vla-policy-results} provides the full protocol, per-suite results, seed-level
consistency checks, and the SmolVLA replication.

Physical evaluation tests whether the same distinctions persist under visual
and geometric changes that are difficult to simulate. The five non-shuffled
variants share 398 training demonstrations across three tasks. Blind, paired
evaluation covers the training distribution, camera blur, and unseen pot
layouts, totaling 28 blocks and 420 trials. Instructions, layouts, and
flow-matching noise are matched within each block.
Appendices~\ref{app:real-robot-setup} and~\ref{app:real-robot-protocol} detail
the training data, block pairing, and evaluation conditions.

\begin{table*}[t]
\centering
\caption{
\textbf{Simulation performance on LIBERO and LIBERO-PRO.}
LIBERO, Position, and PRO mean report mean $\pm$ standard deviation over
three seeds. The remaining perturbation columns report seed means. Retention
is perturbed success divided by standard success.
}
\label{tab:rq3-main}
\footnotesize
\setlength{\tabcolsep}{3pt}
\begin{tabular*}{\textwidth}{@{\extracolsep{\fill}}lcccccccc@{}}
\toprule
& \textbf{LIBERO}
& \multicolumn{6}{c}{\textbf{LIBERO-PRO}}
& \\
\cmidrule(lr){2-2} \cmidrule(lr){3-8}
\textbf{Variant}
& \textbf{mean}
& \textbf{Lang.}
& \textbf{Position}
& \textbf{Object}
& \textbf{Task}
& \textbf{Env.}
& \textbf{mean}
& \textbf{Ret.} \\
\midrule
Baseline (no forecast)
& $92.3{\scriptstyle\,\pm1.3}$
& $86.0$
& $48.8{\scriptstyle\,\pm5.9}$
& $72.3$
& $40.8$
& $49.5$
& $59.5{\scriptstyle\,\pm1.8}$
& $0.635$ \\
Special-token ($M=16$)
& $92.0{\scriptstyle\,\pm1.7}$
& $84.0$
& $45.9{\scriptstyle\,\pm3.2}$
& $79.1$
& $43.2$
& $51.7$
& $60.8{\scriptstyle\,\pm1.5}$
& $0.653$ \\
Special-token ($M=N=256$)
& $90.7{\scriptstyle\,\pm0.8}$
& $85.7$
& $45.3{\scriptstyle\,\pm3.3}$
& $76.6$
& $41.4$
& $52.2$
& $60.2{\scriptstyle\,\pm1.8}$
& $0.658$ \\
Special-token ($M=N=256$)$^{*}$
& $92.6{\scriptstyle\,\pm0.5}$
& $85.8$
& $54.5{\scriptstyle\,\pm2.4}$
& $\mathbf{80.3}$
& $44.3$
& $52.4$
& $63.5{\scriptstyle\,\pm1.1}$
& $0.678$ \\
Vision-token
& $\mathbf{94.7}{\scriptstyle\,\pm1.9}$
& $\mathbf{88.3}$
& $\mathbf{60.9}{\scriptstyle\,\pm1.0}$
& $77.6$
& $\mathbf{46.0}$
& $\mathbf{58.8}$
& $\mathbf{66.3}{\scriptstyle\,\pm0.6}$
& $\mathbf{0.696}$ \\
Vision-token, shuffled future
& $92.2{\scriptstyle\,\pm1.6}$
& $85.0$
& $42.8{\scriptstyle\,\pm6.3}$
& $76.6$
& $38.8$
& $52.7$
& $59.2{\scriptstyle\,\pm1.6}$
& $0.635$ \\
\bottomrule
\end{tabular*}
\vspace{2pt}
\parbox{\textwidth}{\raggedright\footnotesize $^{*}$ Anchored rotary position indices.}
\end{table*}

\begin{figure*}[t]
    \centering
    \includegraphics[width=0.80\linewidth]{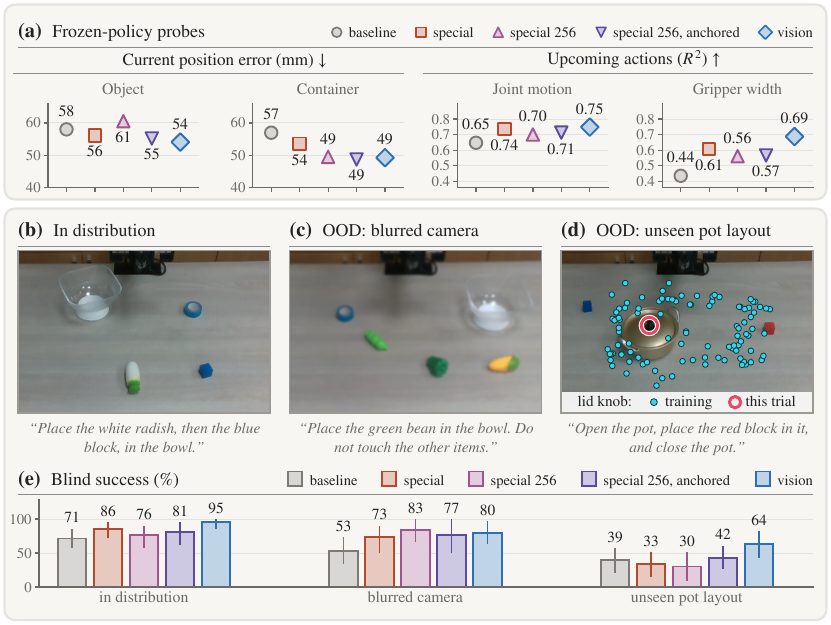}
    \caption{
    \textbf{Vision-token forecasting gives the strongest future-action readout
    and highest pooled real-robot OOD success.}
    \textbf{(a)} Frozen-policy probes on 44 held-out episodes.
    \textbf{(b--d)} Inputs from the training distribution, camera blur, and an
    unseen pot layout. Cyan points mark training lid-grasp positions, and the
    ring marks the trial lid position.
    \textbf{(e)} Success rates with 95\% block-bootstrap intervals.
    }
    \label{fig:rq3-realworld}
\end{figure*}

\paragraph{The robustness advantage persists from simulation to physical manipulation.}
Prediction interfaces separate most clearly under distribution shift
(Table~\ref{tab:rq3-main} and Figure~\ref{fig:rq3-realworld}). In simulation,
all variants retain high standard success, while vision-token forecasting
achieves the strongest object-position and overall PRO performance. On the
real robot, every forecasting variant has a higher success point estimate than
the baseline under camera blur, while vision-token forecasting has the highest
pooled OOD success and the largest unseen-pot improvement. This condition
places the target container roughly twice as far from training layouts as
those layouts lie from one another and requires a rare, unstaged grasp of a
small reflective lid knob (Appendices~\ref{app:real-robot-setup}
and~\ref{app:real-robot-protocol}).

\paragraph{Stable spatial addressing helps, but direct scene-matched forecasting remains strongest.}
In simulation, matching token count and patchwise target granularity provides
little benefit, whereas anchored offsets improve position-shift and overall
perturbed success but remain below vision-token forecasting. On the real
robot, both $M=N=256$ controls improve under camera blur but remain near the
baseline on unseen pot layouts. Simulation therefore provides clearer evidence
for stable spatial addressing, while the physical results support the broader
value of structured predictive supervision. Because the pot condition also
changes the required grasp and multistage geometry, it is a demanding
positional-layout shift rather than a single-axis position intervention.

The shuffled-future control provides the complementary result. Its direct
vision-token route does not improve simulated robustness when the target no
longer matches the scene. An effective route must therefore carry relevant
future change rather than spatial coupling alone.

\paragraph{Visual information about future scene change tracks policy robustness.}
Anchored indices keep forecast gradients spatially aligned and improve
future-displacement readout without similarly improving current-position
decoding (Appendices~\ref{app:vla-m256-delivery}
and~\ref{app:vla-layer-probes}). Across the 12 checkpoints, displacement
readout is associated with position-shift success ($r=0.82$). This association
is consistent with, but does not isolate, a link between routing,
future-change information, and robustness.

Physical-policy probes show the same distinction. Vision-token forecasting
gives the strongest upcoming-action readout, while the two $M=N=256$ controls
improve container-position decoding without matching its unseen-pot performance
(Figure~\ref{fig:rq3-realworld}(a)). Current-position encoding alone therefore
does not explain robustness. Full results appear in
Appendices~\ref{app:vla-layer-probes},
\ref{app:real-robot-results}, and~\ref{app:real-robot-probes}.

\paragraph{VLA robustness improves most when a direct spatial route carries scene-matched future supervision (RQ3).}
At policy scale, the robustness gains from predictive supervision depend on
stable routing and scene-matched targets. Anchoring partially recovers the gap,
whereas shuffling the direct vision-token target removes its gain.

\section{Conclusion}

Future prediction shapes a policy through how its supervision enters and
trains the visual stream. Under matched targets, different interfaces produce
different representations, and routing interventions trace the gap to direct
same-position coupling. This route preserves spatial state, future dynamics,
and action information that attention-mediated routing alone does not recover.
At VLA scale, stable addressing partly closes the robustness gap, while target
shuffling removes direct-route gains. Effective supervision therefore depends
on both a route that shapes the visual representation used for action and
scene-matched future content. Although policy-scale evidence remains
correlational (Appendix~\ref{app:limitations}), it agrees with the controlled
findings across simulated and physical shifts. Future-predictive objectives
should therefore be designed around the policy representations they supervise,
not treated as interchangeable auxiliary heads. Where predictive supervision
enters shapes what a policy learns and how robustly it acts.

\subsection*{AI Use Statement}

Generative AI tools were used to provide feedback on experimental framing and
methodology, refine the conceptual and mathematical presentation, interpret
experimental results, and assist with manuscript organization, drafting, and
language editing. All empirical results reported in the paper were produced by
the authors' training and evaluation pipelines. Generative AI tools were not
used to create experimental measurements or substitute for model training and
evaluation. The authors reviewed all AI-assisted material and take
responsibility for the final content of this work, including text, claims, and
artifacts produced with the aid of generative AI.

\subsection*{Reproducibility Statement}

We will provide the code used for controlled-model training, VLA training,
routing interventions, representation probes, evaluation, and figure
generation. Appendices~\ref{app:controlled-representation-future-prediction}
and~\ref{app:gradient-mechanism} document the controlled architectures,
training protocols, probe definitions, and routing interventions.
Appendices~\ref{app:vla-scale-evaluation}
and~\ref{app:real-robot-evaluation} specify the VLA variants, data splits,
evaluation protocols, aggregation procedures, and uncertainty estimates.
Appendix~\ref{app:derivation-gradients} provides the formal derivation of the
prediction credit routes. VLA policies were trained on
NVIDIA RTX PRO 6000 GPUs, while the controlled models and other auxiliary
training runs used NVIDIA RTX 3090 GPUs.

\bibliography{iclr2027_conference}
\bibliographystyle{iclr2027_conference}

\clearpage
\appendix

\counterwithin{table}{section}
\counterwithin{figure}{section}
\counterwithin{equation}{section}

\addtocontents{toc}{\protect\setcounter{tocdepth}{2}}
\setcounter{tocdepth}{2}
\renewcommand{\contentsname}{Supplementary Material}
\tableofcontents
\clearpage

\section{Predictive-Supervision Interfaces and VLA Robustness}
\label{app:prior-interfaces}

\subsection{Interface Taxonomy}

Future-predictive robot policies differ in what they predict and in where the
prediction objective enters the trainable policy.
Table~\ref{tab:future-token-interfaces} organizes representative methods by
their prediction readout and by how the resulting supervision reaches the
policy's visual stream.

\begin{table}[!htbp]
\caption{
Representative future-prediction interfaces and how their supervision reaches
trainable policy vision tokens.
}
\label{tab:future-token-interfaces}
\centering
\small
\setlength{\tabcolsep}{3.5pt}
{\renewcommand{\arraystretch}{1.12}
\begin{tabularx}{\linewidth}{@{}>{\raggedright\arraybackslash}p{0.23\linewidth}>{\raggedright\arraybackslash}p{0.35\linewidth}>{\raggedright\arraybackslash}X@{}}
\toprule
\textbf{Prediction readout}
&
\textbf{Representative methods}
&
\textbf{How supervision reaches policy vision tokens}
\\
\midrule

Autoregressive output tokens
&
CoT-VLA~\citep{Zhao2025-br}
&
Sequence-mediated, no explicit same-patch attachment
\\

Learned/query tokens
&
FLARE, DreamVLA, WoG, HiF-VLA~\citep{Zheng2025-du,Zhang2025-yd,
Su2026-worldguidance,Lin2025-hifvla}
&
Attention-mediated, no explicit same-patch attachment
\\

Predictive latent alignment
&
VLA-JEPA, FutureVLA~\citep{Sun2026-fb,Xu2026-futurevla}
&
Latent-state or intermediate alignment, no explicit same-patch attachment
\\

Future readout + spatial auxiliary
&
FoMoVLA~\citep{Li2026-fomovla}
&
Indirect forecast readout with a direct spatial auxiliary
\\

Separate world model
&
AHEAD~\citep{Syed2026-ahead}
&
Patch-aligned outside the frozen policy
\\

\midrule
Special tokens
&
This study
&
Attention-mediated, no guaranteed same-position path
\\

Policy vision tokens
&
This study
&
Direct, same-position forecast path
\\

\bottomrule
\end{tabularx}
}
\end{table}
\FloatBarrier

These categories separate where a prediction is read out from how its loss
reaches the policy's visual stream. A learned or query-based readout may acquire
spatial structure without providing a direct same-position path. A direct
spatial auxiliary may also coexist with an indirect future readout. We
therefore use \emph{direct coupling} to mean an explicit same-position
computational path from prediction $i$ to policy visual token $i$, rather than
a one-to-one target assignment or an attention pattern alone.
Appendix~\ref{app:derivation-gradients} formalizes this distinction.

\subsection{Evaluating Robustness in VLA Policies}

Standard task success does not fully characterize how a VLA policy behaves
beyond the configurations encountered during training. Recent benchmarks
therefore evaluate policies under controlled changes to objects, spatial
arrangements, language, and environments. LIBERO-PRO and LIBERO-Plus organize
such perturbations around established manipulation
tasks~\citep{Zhou2026-gm, Fei2025-jb}, while Colosseum V2 broadens evaluation
across visual, semantic, and physical variations~\citep{Morgan2026-zf}.
Together, these benchmarks distinguish in-distribution task completion from
robustness under distribution shift. They primarily characterize behavioral
outcomes, however, leaving how training objectives shape the representations
underlying robust behavior less understood.

\subsection{Relation to Our Controlled Comparison}

The interfaces in Table~\ref{tab:future-token-interfaces} vary in their
targets, horizons, backbones, and training stages as well as in how predictive
supervision reaches the visual stream. RQ1 isolates supervision placement by
holding the future target, readout count, backbone, and optimization fixed
(Appendix~\ref{app:controlled-representation-future-prediction}). RQ2
separates spatial address from direct same-position coupling
(Appendix~\ref{app:gradient-mechanism}). RQ3 then examines whether routing and
scene-matched future content shape VLA robustness in simulation and physical
manipulation (Appendices~\ref{app:vla-scale-evaluation}
and~\ref{app:real-robot-evaluation}).

\section{Controlled Forecast and Representation Analysis (RQ1)}
\label{app:controlled-representation-future-prediction}

This appendix provides the full setup and evaluation details for RQ1. The
comparison holds the target construction, prediction horizon, number of
forecast readouts, backbone, and optimization fixed, while changing the
token stream from which each patch forecast is read.

\subsection{Matched Interfaces and Training}
\label{app:controlled-forecasting-setting}

\begin{figure}[!htbp]
    \centering
    \includegraphics[width=\linewidth]{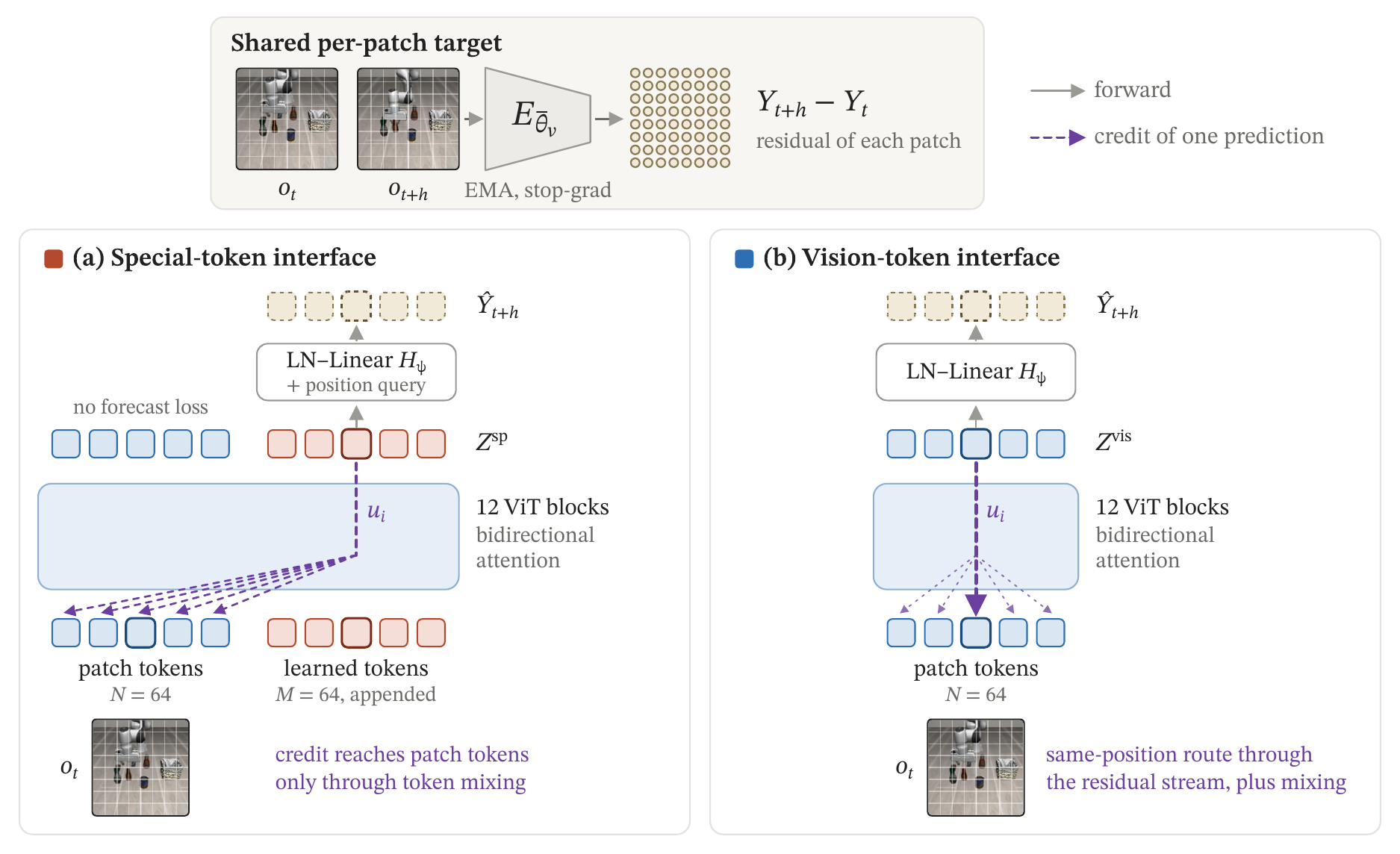}
    \caption{
    \textbf{Matched prediction interfaces used in RQ1.}
    \textbf{(a)} The special-token interface predicts from one added token per
    future patch, so forecast errors reach the visual stream through contextual
    mixing.
    \textbf{(b)} The vision-token interface predicts from the corresponding
    patch token, retaining contextual mixing and adding a direct same-position
    route. Dashed arrows show the backward paths.
    }
    \label{fig:controlled-forecasting-interfaces}
\end{figure}

Both interfaces encode the same $8\times8$ grid of current-image patches and
predict one residual for each of the 64 future patches
(Figure~\ref{fig:controlled-forecasting-interfaces}). The special-token
interface appends $M=N=64$ tokens and predicts target $i$ from contextualized
special token $\tilde{s}_i$. A learned position query identifies the
corresponding forecast readout after contextualization, but does not impose a
same-position attention route to visual patch $i$. The vision-token interface
instead predicts from contextualized patch token $\tilde{v}_i$, giving each
forecast error a direct route to the visual position it describes. Both
interfaces use a token-wise LayerNorm followed by a linear prediction layer.
Adding the special tokens increases the encoder sequence from 64 to 128 tokens,
which is part of the interface construction examined here. Each model maintains
its own momentum encoder, so matching refers to the target definition and
horizon rather than numerically identical target features throughout training.

\begin{table}[!htbp]
\centering
\small
\caption{\textbf{Controlled RQ1 training setup.} All settings are shared by
the special-token and vision-token interfaces.}
\label{tab:controlled-training-setup}
\begin{tabularx}{\linewidth}{@{}lX@{}}
\toprule
\textbf{Component} & \textbf{Setting} \\
\midrule
Data &
500 \textsc{LIBERO-Object} episodes from \texttt{agentview\_rgb}, with every
tenth episode held out, giving 450 training and 50 validation episodes \\
Training clips &
Every valid clip start frame in the training episodes \\
Input &
$128\times128$ images, $16\times16$ patches, and an $8\times8$ grid of
64 visual tokens \\
Encoder &
12-block ViT with width 192, three attention heads, and learned absolute
positional embeddings \\
Future target &
Stop-gradient patch residual at $h=24$ from a momentum encoder with
$m=0.996$ \\
Loss &
Per-patch squared error weighted by the normalized norm of each target
residual \\
Optimization &
AdamW for 4,000 steps, batch size 128, learning rate $10^{-3}$, 200 warmup
steps followed by cosine decay, and weight decay $0.05$ \\
Supervision &
Future prediction only, with no action or language objective \\
Seeds &
Five training seeds \\
\bottomrule
\end{tabularx}
\end{table}

\FloatBarrier

\subsection{Decoded Forecast Evaluation}
\label{app:decoded-forecast-evaluation}

For each frozen encoder, we fit a pixel decoder using only features from real
frames. Forecast features are excluded from decoder training, so decoded
forecast quality reflects how the predicted residual moves a real-frame
representation rather than how well the decoder adapts to forecast outputs.
Table~\ref{tab:decoded-forecast-setup} summarizes the decoder training and
evaluation protocol.

\begin{table}[!htbp]
\centering
\small
\caption{\textbf{Decoded-forecast evaluation setup.}}
\label{tab:decoded-forecast-setup}
\begin{tabularx}{\linewidth}{@{}lX@{}}
\toprule
\textbf{Component} & \textbf{Setting} \\
\midrule
Decoder data &
35 held-out episodes for fitting and 15 disjoint episodes for evaluation \\
Decoder input &
Features from real current and future frames only \\
Optimization &
Adam for 12,000 steps, batch size 48, learning rate
$1.5\times10^{-3}$ with cosine decay \\
Objective &
$\ell_1+0.5\,\mathrm{MSE}$ \\
Evaluation set &
617 samples per seed, comprising 418 near-center and 199 spatial-tail samples \\
Moving pixels &
Mean absolute RGB change above $0.08$ for images scaled to $[0,1]$,
excluding samples with at most 20 moving pixels \\
\bottomrule
\end{tabularx}
\end{table}

To evaluate forecasts in pixel space, we apply the same fitted decoder $D$ to
the current spatial features $z_t$, the predicted future features
$z_t+\hat r_t$, and the true future features $z_{t+h}$. This gives
\begin{equation}
\hat o_t=D(z_t), \qquad
\hat o_{t+h}^{\mathrm{pred}}=D(z_t+\hat r_t), \qquad
\hat o_{t+h}^{\mathrm{oracle}}=D(z_{t+h}).
\label{eq:app-decoded-forecasts}
\end{equation}
The decoded current features provide the copy baseline, while the decoded true
future features provide an oracle for how much future information can be
recovered by the fitted decoder. The oracle is therefore not a perfect
pixel-level prediction.

With all errors measured against the true future frame, these two references
define forecast gap closure as
\begin{equation}
G=
\frac{\mathrm{MSE}_{\mathrm{copy}}-\mathrm{MSE}_{\mathrm{pred}}}
     {\mathrm{MSE}_{\mathrm{copy}}-\mathrm{MSE}_{\mathrm{oracle}}}.
\label{eq:app-gap-closure}
\end{equation}
Copying the current reconstruction gives $G=0$, while decoding the true future
feature gives $G=1$. The moving-pixel results use the criterion summarized in
Table~\ref{tab:decoded-forecast-setup}.

Beyond gap closure, we measure whether each forecast captures the amount and
location of scene change. Decoded change magnitude is the spatial mean of
$\lVert D(z_t+\hat r_t)-D(z_t)\rVert_2$, while actual change magnitude is
the spatial mean of $\lVert o_{t+h}-o_t\rVert_2$. Decoded-change
slope is the sample-wise slope between these quantities. Oracle-change slope
uses $D(z_{t+h})$ in place of the predicted future features. Change-map
correlation is the pixel-wise Pearson correlation between decoded and actual
change maps.

\begin{table}[!htbp]
\centering
\small
\caption{
\textbf{Decoded-forecast results.}
Gap closure is reported in percent. Gap closure and change slopes report mean
$\pm$ standard deviation over five seeds. Change-map correlation reports the
pooled median, with brackets giving the range of seed-level medians.
}
\label{tab:decoded-gap-closure}

\begin{tabular}{@{}lrrrr@{}}
\toprule
& \multicolumn{2}{c}{\textbf{All pixels}} &
  \multicolumn{2}{c}{\textbf{Moving pixels}} \\
\cmidrule(lr){2-3}\cmidrule(l){4-5}
\textbf{Split} &
\textbf{Vision} & \textbf{Special} &
\textbf{Vision} & \textbf{Special} \\
\midrule
All          & $45.0\pm1.2$ & $0.3\pm2.9$  & $56.4\pm1.6$ & $3.0\pm2.2$ \\
Near-center  & $42.8\pm1.4$ & $-0.1\pm3.6$ & $54.4\pm1.7$ & $2.9\pm2.0$ \\
Spatial tail & $49.0\pm1.3$ & $1.1\pm2.0$  & $60.7\pm1.4$ & $3.1\pm2.9$ \\
\bottomrule
\end{tabular}

\vspace{6pt}

\begin{tabularx}{\linewidth}{@{}Xrr@{}}
\toprule
\textbf{Forecast diagnostic} & \textbf{Vision} & \textbf{Special} \\
\midrule
Decoded-change slope &
$0.535\pm0.035$ & $0.013\pm0.044$ \\
Oracle-change slope &
$0.925\pm0.012$ & $1.014\pm0.017$ \\
Change-map correlation &
$0.709\,[0.703,0.717]$ & $0.324\,[0.269,0.368]$ \\
\bottomrule
\end{tabularx}
\end{table}

Vision-token forecasts capture both the magnitude and spatial location of
future change, whereas special-token forecasts remain close to the current
frame. The comparable oracle slopes show that this difference is not caused by
a weaker pixel decoder for the special-token model. Figure~\ref{fig:rq1-forecast-probes}(a) of the main
paper complements these aggregate metrics with illustrative forecasts in
which scene change is clearly visible. Within each split, we retain the half
of samples with the largest moving area and choose the example closest to the split medians of gap closure
and spatial agreement. The spatial-tail example contains more object transport
and larger changes, so its higher gap closure should not be interpreted as
stronger spatial-tail forecasting.

\FloatBarrier

\subsection{Frozen-Feature Probes}
\label{app:frozen-feature-probes}

The frozen-feature evaluation uses a common bank of 2,000 clips from
the 50 validation episodes and forms five task-stratified episode partitions,
each holding out one quarter of the episodes. We fit the same ridge probes to
the visual-token grids of both encoders. For the special-token model, we
additionally probe its patch-ordered special-token stream to determine where
the measured information remains linearly accessible.

\noindent\textbf{Position.}
We average-pool the token grid to $4\times4$ and regress the current object
centers.

\noindent\textbf{Identity.}
We predict the target identity, among ten objects, from the token at the
object's patch.

\noindent\textbf{Future displacement.}
For each visible object that moves by more than one quarter of a patch, we
regress its displacement $h=24$ frames into the future from its current patch
token.

\noindent\textbf{Action chunk.}
We regress the upcoming 24-step action sequence from the pooled token grid.

\noindent\textbf{Geometry.}
At every token position, we predict the vector from that token to each object
center. This object-vector field measures whether object-centered geometry
remains accessible across the visual grid.

Identity is evaluated by classification accuracy. All regression probes use
$R^2$ relative to the training-set mean. The ridge strength is 10 for position,
identity, and action chunk, and 1 for future displacement and geometry.

\noindent\textbf{Spatial-transfer split.}
A clip belongs to the near-center split when the target object lies within the
median distance of its task-specific mean position, and to the spatial-tail
split otherwise. Each probe is fitted on near-center clips from the training
episodes and evaluated on both splits from the held-out episodes. Results are
averaged over the five partitions. Depending on the partition, fitting uses
between 1,077 and 1,126 clips, while evaluation uses between 256 and 305
near-center clips and between 113 and 150 spatial-tail clips. All ten tasks are
represented. The encoder itself was trained on the full training episodes, so
this measures probe-level spatial transfer rather than an unseen distribution
for the encoder. The main figure distinguishes the two splits by outline style.

\noindent\textbf{Effective rank.}
We measure representation breadth as
$\exp(-\sum_k p_k\log p_k)$, where
$p_k=\sigma_k^2/\sum_j\sigma_j^2$ is the normalized squared singular-value
spectrum of the mean-centered token features. All controlled-model ranks use
the same 8,192 token embeddings from 64 tokens across 128 validation clips.
Effective ranks computed from different sample sizes are not directly comparable.

\begin{table}[!htbp]
\centering
\scriptsize
\setlength{\tabcolsep}{4pt}
\caption{Frozen-feature probes underlying Figure 2(d) of the main paper.
Values are mean $\pm$ standard deviation over five seeds. Identity is
classification accuracy, and all other targets use $R^2$. Near and Tail denote
the near-center and spatial-tail splits. The two special-token columns
read the visual and special-token grids of the same encoder.}
\label{tab:controlled-probes}
\begin{tabular}{@{}llrrr@{}}
\toprule
\textbf{Target} & \textbf{Split} & \textbf{Vision-token visual} &
\textbf{Special-token visual} & \textbf{Special-token stream} \\
\midrule
Position & Near & $0.939{\scriptstyle\pm0.005}$ & $0.577{\scriptstyle\pm0.193}$ & $0.508{\scriptstyle\pm0.294}$ \\
& Tail & $0.574{\scriptstyle\pm0.025}$ & $0.138{\scriptstyle\pm0.134}$ & $0.131{\scriptstyle\pm0.207}$ \\
Identity & Near & $0.931{\scriptstyle\pm0.024}$ & $0.649{\scriptstyle\pm0.128}$ & $0.494{\scriptstyle\pm0.171}$ \\
& Tail & $0.304{\scriptstyle\pm0.030}$ & $0.182{\scriptstyle\pm0.043}$ & $0.138{\scriptstyle\pm0.044}$ \\
Future displacement & Near & $0.796{\scriptstyle\pm0.020}$ & $0.529{\scriptstyle\pm0.121}$ & $0.460{\scriptstyle\pm0.193}$ \\
& Tail & $0.450{\scriptstyle\pm0.117}$ & $0.332{\scriptstyle\pm0.100}$ & $0.304{\scriptstyle\pm0.167}$ \\
Action chunk & Near & $0.631{\scriptstyle\pm0.022}$ & $0.351{\scriptstyle\pm0.190}$ & $0.342{\scriptstyle\pm0.224}$ \\
& Tail & $0.612{\scriptstyle\pm0.031}$ & $-0.031{\scriptstyle\pm0.401}$ & $-0.185{\scriptstyle\pm0.785}$ \\
Geometry & Near & $0.527{\scriptstyle\pm0.018}$ & $0.556{\scriptstyle\pm0.102}$ & $0.445{\scriptstyle\pm0.153}$ \\
& Tail & $0.411{\scriptstyle\pm0.046}$ & $0.235{\scriptstyle\pm0.443}$ & $0.229{\scriptstyle\pm0.276}$ \\
\bottomrule
\end{tabular}
\end{table}

Table~\ref{tab:controlled-probes} gives the exact values behind the main
paper's probe profile. Vision-token features retain the strongest spatial-tail
position, future-displacement, action, and geometry readouts. Neither readout
site in the special-token model recovers this profile. Identity follows the same ordering but degrades for every model, indicating
that the interface advantage is strongest for spatial and action-related
structure rather than uniform across representation properties. The same contrast appears in representation breadth.
Vision-token features have effective rank $36.0\pm0.5$, compared with
$6.3\pm5.1$ for the visual tokens of the special-token encoder and
$7.4\pm5.7$ for its special-token stream.

\FloatBarrier

\subsection{Frozen-Teacher and Positional-Structure Controls}
\label{app:frozen-teacher-control}

The primary RQ1 comparison uses a momentum target that co-evolves with each
student. We test whether the special-token result is specific to this target
construction by replacing it with two frozen teachers. The first teacher is a
separately trained vision-token encoder from the same controlled forecasting
recipe. It preserves the forecast-trained feature space while removing target
co-evolution. The second is frozen SigLIP-B/16. For this condition, input frames
are resized to $224\times224$, the $14\times14$ final hidden grid is area-pooled
to $8\times8$, and a fixed PCA projection maps the 768-dimensional features to
192 dimensions. The projection is fitted on 8,192 training frames, preserves
$92.1\%$ of the feature variance, and is normalized to unit mean variance.

All other controlled-model settings, including the prediction horizon,
motion-weighted objective, optimization, data split, and five training seeds,
match Appendix~\ref{app:controlled-forecasting-setting}. Under the frozen
SigLIP target, both interfaces are decoded in the same teacher feature space
with one shared pixel decoder. Under the frozen forecast-trained target, each
teacher uses the pixel decoder already fitted to its feature space. We also
repeat the momentum and frozen-SigLIP comparisons with 1-D rotary positions.
In this construction, special token $i$ is placed at index $N+i$, providing a
stable relative offset to visual patch $i$ without adding the direct visual
residual path of the vision-token interface.

\begin{figure}[!htbp]
    \centering
    \includegraphics[width=\linewidth]{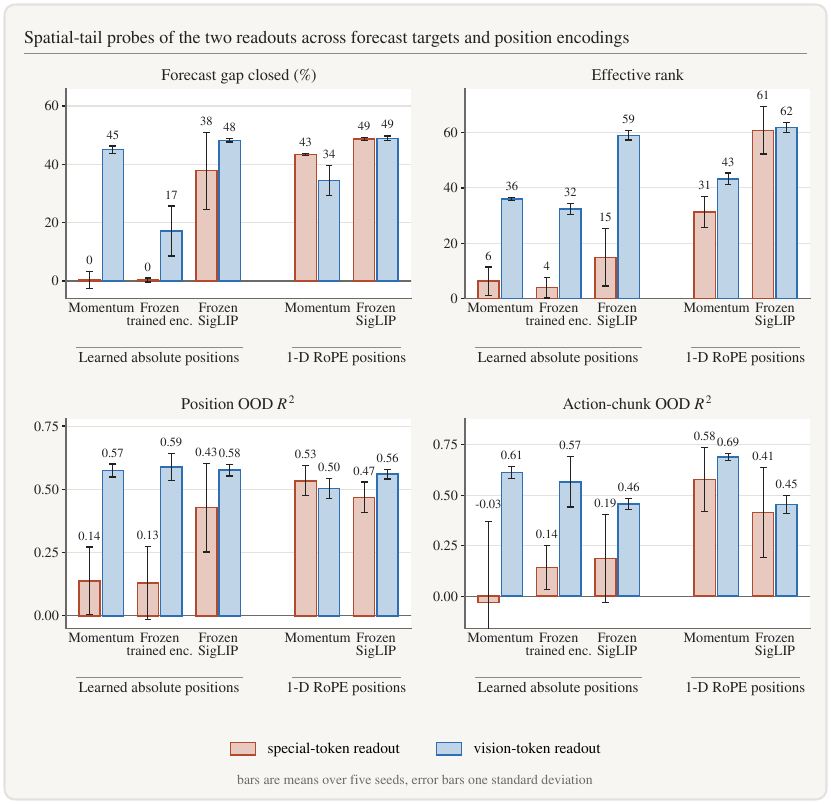}
    \caption{
    \textbf{Forecast targets and positional structure jointly determine the
    special-token outcome.}
    Learned-absolute-position models use the momentum target from RQ1, a frozen
    forecast-trained encoder, or frozen SigLIP features. The 1-D RoPE models use
    either the momentum or frozen-SigLIP target with a stable
    special-token-to-patch offset. Bars show means over five seeds and error
    bars show one standard deviation. Forecast gap closure, effective rank,
    and spatial-tail position and action probes are reported. Rank and probes
    read the visual-token representation learned by the corresponding
    prediction interface. Forecast gap closure should be compared between
    interfaces within a target condition, because decoder feature spaces differ
    across target families.
    }
    \label{fig:app-frozen-teacher-control}
\end{figure}

\paragraph{Freezing a forecast-trained teacher does not rescue the unindexed
special-token interface.}
With learned absolute positions, replacing the momentum target by a frozen
copy of a separately trained forecasting encoder leaves the special-token
forecast near the copy baseline: gap closure is $0.3\%$, compared with
$17.1\%$ for vision-token readout. Its visual representation also remains
narrow, with effective rank $3.9$ versus $32.3$, and transfers weakly on
position ($R^2=0.13$ versus $0.59$) and action chunks ($0.14$ versus $0.57$).
These results closely reproduce the momentum-target ordering while removing
target co-evolution. The failure is therefore not explained solely by a
self-referential momentum target.

\paragraph{A frozen pretrained target improves forecasting without closing
the representation gap.}
With learned absolute positions and frozen SigLIP features, special-token gap
closure rises to $37.8\%$, compared with $48.2\%$ for vision-token readout.
Thus, a fixed semantic target allows the special-token model to produce a
substantial decoded forecast. Its visual stream nevertheless remains
lower-rank ($14.9$ versus $59.1$) and weaker under spatial-tail evaluation for
position ($0.43$ versus $0.58$) and action chunks ($0.19$ versus $0.46$).
The special-token runs are also less stable: two of five seeds have effective
rank below five. The interface ordering is not uniform across every property;
for example, future-displacement readout under the frozen SigLIP target is not
separated reliably. The result supports the narrower conclusion that forecast
quality alone does not determine whether the learned visual representation
transfers.

\paragraph{Stable positional addressing narrows the interface gap.}
The 1-D RoPE construction substantially changes the special-token result. With
the momentum target, special-token readout closes more of the decoded forecast
gap than the vision-token reference ($43.5\%$ versus $34.5\%$), while retaining
a lower action-chunk readout ($0.58$ versus $0.69$). With frozen SigLIP, the two
interfaces are nearly matched in forecast gap closure ($48.6\%$ versus
$48.9\%$), effective rank ($60.9$ versus $61.9$), and action transfer ($0.41$
versus $0.45$), although vision-token features retain the higher position
readout ($0.56$ versus $0.47$). A frozen informative target and stable spatial
address can therefore make an attention-mediated special-token interface
competitive. The main RQ1 result should be read as evidence about matched
unindexed interfaces under the momentum target, while the broader target sweep
shows that prediction target and positional structure interact with where
supervision is attached.

\FloatBarrier

\section{Credit-Routing Interventions (RQ2)}
\label{app:gradient-mechanism}

This appendix provides the measurement definitions, intervention protocols,
and complete results supporting RQ2. The analyses separate two properties of
predictive supervision. One is whether forecast gradients reach the matching
visual region, and the other is whether each forecast has a direct same-position
path to that region. Appendix~\ref{app:derivation-gradients} provides the
corresponding gradient derivation.

\subsection{Measuring Spatial Alignment of Forecast Credit}
\label{app:gradient-delivery}

We measure spatial alignment by asking where the gradient from each regional
forecast loss reaches the visual-token grid. We divide the forecast targets and
visual tokens into the same $4\times4$ grid. For each forecast region, we
isolate its loss and compute the fraction of the resulting visual-token
gradient norm that falls within the corresponding image region. We call the
average over the 16 regions the \emph{same-region gradient share}. If the
gradient destination is independent of the forecast region, the expected share
is $1/16=0.0625$.

To distinguish spatial alignment from the diversity of the delivered updates,
we also report centered gradient rank. For each sample, let
$G\in\mathbb{R}^{N\times d}$ stack the gradient of the full forecast loss
with respect to the $N$ visual-token activations. We subtract the component
shared across patch positions,
$G_{\mathrm{c}}=(I-N^{-1}\mathbf{1}\mathbf{1}^\top)G$, and compute
\[
\operatorname{rank}_{\mathrm{eff}}(G_{\mathrm{c}})
=\exp\!\left(-\sum_k p_k\log p_k\right),
\qquad
p_k=\frac{\sigma_k^2}{\sum_l\sigma_l^2},
\]
where $\sigma_k$ are the singular values of $G_{\mathrm{c}}$. We average
this quantity over samples. Higher values indicate a broader set of
position-dependent update directions. We apply both diagnostics to the
controlled encoders and trained $\pi_{0.5}$ policies to test whether the
interface-level routing difference persists at VLA scale.

In the controlled model, measurements are taken at the input to the first
Transformer block, the earliest point shared by both interfaces. We use the
untrained initialization to isolate the delivery structure before either model
learns a routing pattern. The shuffled-route reference permutes where the
vision-token gradient arrives while preserving its values. It differs from the
VLA shuffled-target control, which preserves the vision-token route but changes
the future target.

In $\pi_{0.5}$, measurements are taken from the prefix input embeddings after
the SigLIP projection and before the language model. We use 400
training-demonstration samples balanced across four LIBERO suites.

\begin{table}[htbp]
\centering
\small
\caption{Spatial alignment of forecast credit. Values are mean $\pm$
standard deviation. Controlled-model values use five initializations.
Vision-token and shuffled-target VLA values use 400 samples from one available
checkpoint. The special-token VLA row pools 1200 samples across three
checkpoints.}
\label{tab:gradient-delivery}
\begin{tabular}{@{}lrr@{}}
\toprule
\textbf{Interface} &
\shortstack{\textbf{Same-region}\\\textbf{gradient share}} &
\shortstack{\textbf{Centered}\\\textbf{gradient rank}} \\
\midrule
\multicolumn{3}{@{}l}{\emph{Controlled, initialization}} \\
Vision-token & $0.673\pm0.041$ & $5.6\pm0.4$ \\
Shuffled route & $0.062\pm0.002$ & $5.8\pm0.4$ \\
Special-token & $0.062\pm0.000$ & $1.8\pm0.1$ \\
\midrule
\multicolumn{3}{@{}l}{\emph{$\pi_{0.5}$, trained}} \\
Vision-token & $0.694\pm0.070$ & $26.8\pm8.5$ \\
Shuffled target & $0.701\pm0.034$ & $23.6\pm7.3$ \\
Special-token & $0.142\pm0.048$ & $18.5\pm5.9$ \\
\bottomrule
\end{tabular}
\end{table}

At initialization, vision-token forecast credit is concentrated in the
corresponding image region, whereas the special-token and shuffled routes
remain at chance. The routing difference therefore exists before either
interface has learned a spatial correspondence. Centered gradient rank provides
a secondary distinction, but the clearest separation is where forecast credit
reaches the visual stream.

\begin{figure}[t]
    \centering
    \includegraphics[width=0.9\linewidth]{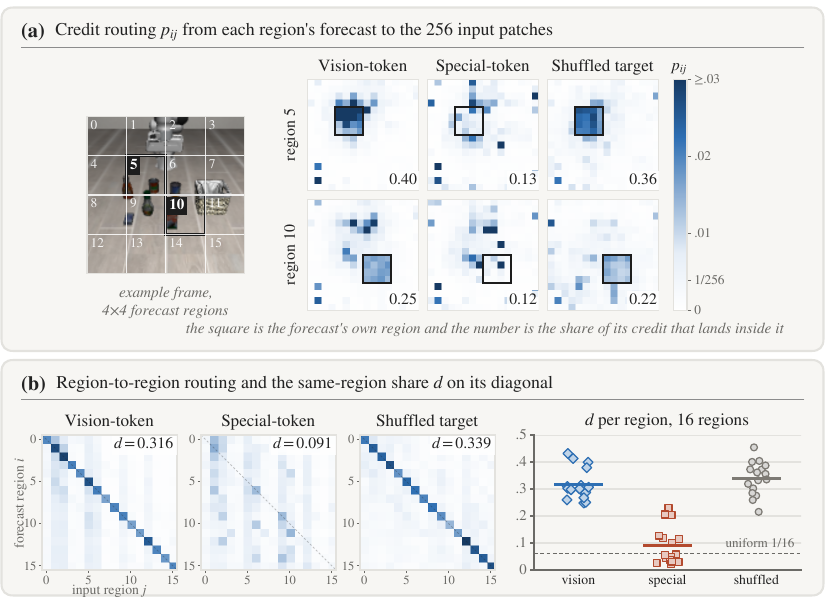}
    \caption{
    \textbf{Spatial routing of forecast credit in $\pi_{0.5}$ (LIBERO-Object).}
    Forecasts are pooled to the $4\times4$ region grid of the special-token
    interface, so all conditions produce 16 forecasts over the same 256 input
    patches. \textbf{(a)} Row-normalized routing $p_{ij}$ over the input
    patches for two forecast regions in each condition. The square marks the
    corresponding input region, and the number gives the share of credit that
    reaches it. \textbf{(b)} Region-to-region routing with forecast regions
    as rows and input regions as columns. The rightmost plot compares the
    same-region share $d$ across all 16 regions with the uniform level $1/16$.
    }
    \label{fig:app-vla-routing}
\end{figure}

\paragraph{Applying the same routing analysis at VLA scale.}
We repeat the controlled spatial-routing measurement on trained $\pi_{0.5}$
policies. For each forecast region, we isolate its loss and measure where the
resulting gradient reaches the visual-token grid. We also retain the finer
row-normalized routing matrix
$p_{ij}\propto\lVert\partial\ell_i/\partial v_j\rVert_2$ for forecast
region $i$ and visual patch $j$. Figure~\ref{fig:app-vla-routing} shows the
patch-level maps and their region-to-region aggregation.

The VLA maps reproduce the routing distinction observed in the controlled
model. Vision-token forecasting remains diagonally concentrated, and the same
pattern persists when the future target is shuffled. Special-token forecasting
instead returns credit to similar object regions across different forecast
locations. The similarity between the vision-token and shuffled-target
conditions shows that the prediction interface continues to determine the
spatial destination of forecast credit after joint prediction and action
training.

Having established that the same routing distinction persists at VLA scale,
we return to the controlled model to isolate the role of direct same-position
coupling.

\subsection{Aligned-Path Construction and Sensitivity}
\label{app:aligned-path-rescue}

Figure~\ref{fig:rq2-credit}(a) shows that a moderate aligned path moves the special-token model
away from the collapsed solution. This subsection provides
the exact construction, characterizes sensitivity to coupling strength, and
removes the path after training to distinguish a learned representation change
from direct forward access.

The intervention adds the corresponding final visual token to each
special-token readout after encoder normalization:
\[
z_i^{(\gamma)}=\tilde s_i+p_i+\gamma\tilde v_i,
\qquad
\hat r_i=H_\psi\!\left(z_i^{(\gamma)}\right).
\]
Here, $\tilde s_i$ and $\tilde v_i$ are the contextualized special and visual
tokens, and $p_i$ is the learned position query. Because $M=N=64$, each
forecast readout has one corresponding visual position. Setting $\gamma=0$
recovers the original special-token interface, while positive $\gamma$
introduces a path used by both forward prediction and the gradient of the
forecasting loss. All other data, targets, model components, and optimization
settings match Appendix~\ref{app:controlled-forecasting-setting}. Each
condition uses five training seeds.

\paragraph{Credit-map computation.}
The credit maps in Figure~\ref{fig:rq2-credit}(a) isolate the forecast loss of one target
region containing $2\times2$ patches and measure its gradient norm at every
patch before the first Transformer block. Each map averages 256 clips and five initializations, and
all maps use the same square-root color scale. The outlined region identifies
the source of the forecast loss. Values printed below the maps report the
same-region gradient share averaged over all 16 source regions rather than the
single region shown, as defined in Appendix~\ref{app:gradient-delivery}.

\begin{table}[htbp]
\centering
\scriptsize
\caption{Aligned-path rescue in the controlled special-token model. Values are
mean $\pm$ standard deviation over five seeds. Init. share denotes the
same-region gradient share at initialization. \emph{Path off} evaluates the
trained model after setting $\gamma=0$ without further optimization.}
\label{tab:aligned-path-rescue}
\begin{tabular}{@{}lrrrrr@{}}
\toprule
\textbf{Condition} & \textbf{Init. share} & \textbf{Eff. rank} &
\shortstack{\textbf{Position $R^2$}\\\textbf{(spatial tail)}} &
\shortstack{\textbf{Gap closed}\\\textbf{(path on, \%)}} &
\shortstack{\textbf{Gap closed}\\\textbf{(path off, \%)}} \\
\midrule
Special-token ($\gamma=0$) & $0.062\pm0.000$ & $6.3\pm5.2$ &
$0.14\pm0.13$ & $0.3\pm2.9$ & -- \\
$\gamma=0.05$ & $0.071\pm0.004$ & $9.7\pm12.3$ &
$0.21\pm0.24$ & $-4.3\pm35.5$ & $0.4\pm12.9$ \\
$\gamma=0.2$ & $0.149\pm0.023$ & $46.1\pm4.5$ &
$0.52\pm0.07$ & $40.3\pm2.7$ & $19.7\pm3.4$ \\
$\gamma=1$ & $0.484\pm0.056$ & $50.9\pm9.3$ &
$0.50\pm0.11$ & $42.4\pm1.9$ & $16.4\pm5.4$ \\
Vision-token & $0.673\pm0.041$ & $36.0\pm0.5$ &
$0.57\pm0.03$ & $45.0\pm1.2$ & -- \\
\bottomrule
\end{tabular}
\end{table}

\paragraph{Sensitivity to coupling strength.}
The sweep reveals a transition rather than a monotonic benefit from stronger
coupling. A weak path remains unstable across seeds, while moderate coupling
consistently moves the model away from the collapsed special-token solution
and recovers forecast quality, representation breadth, and spatial transfer.
Increasing the path further does not improve transfer uniformly, indicating
that the result is not explained by greater spatial alignment alone.

\paragraph{The recovery is not only a readout shortcut.}
Because the added visual token is also available during the forward pass, the
improved forecast could reflect direct feature access rather than a change in
what the special-token pathway learns. We therefore set $\gamma=0$ after
training without updating either the model or decoder. The decoded forecast
becomes weaker but does not return to the original special-token solution. The
remaining recovery shows that training changes what the special-token pathway
retains rather than relying entirely on direct forward access. Together with
the recovered visual-token probes in
Table~\ref{tab:aligned-path-full-probes}, this indicates that the intervention
also changes the learned visual representation. Stronger coupling does not
preserve more of the recovery after removal, suggesting that it primarily
increases dependence on the added path.

\begin{figure}[t]
    \centering
    \includegraphics[width=\linewidth]{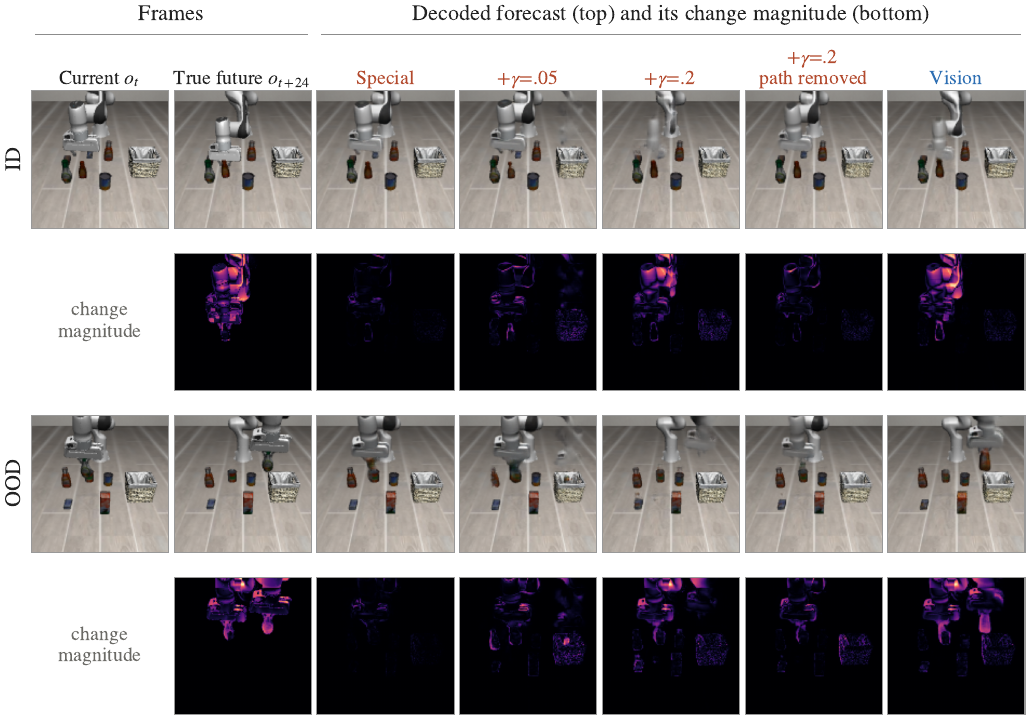}
    \caption{
    \textbf{Decoded forecasts under aligned-path rescue.}
    Rows reuse the near-center and spatial-tail examples from the RQ1 figure,
    and each condition uses its median gap-closure seed. The moderate aligned
    path predicts the future arm motion. Setting $\gamma=0$ after training
    leaves a weaker but visible change through the special-token pathway
    without retraining the model or decoder. Change maps use a common scale
    within each row.
    }
    \label{fig:app-aligned-path-decoded}
\end{figure}

\begin{table}[htbp]
\centering
\scriptsize
\caption{Additional spatial-tail frozen-feature probes for the aligned-path
sensitivity analysis. All readouts use visual tokens and values are mean
$\pm$ standard deviation over five seeds.}
\label{tab:aligned-path-full-probes}
\begin{tabular}{@{}lrrrr@{}}
\toprule
\textbf{Condition} & \textbf{Identity} & \textbf{Future disp.} &
\textbf{Action chunk} & \textbf{Geometry} \\
\midrule
Special-token ($\gamma=0$) & $0.18\pm0.04$ & $0.33\pm0.10$ &
$-0.03\pm0.40$ & $0.24\pm0.44$ \\
$\gamma=0.05$ & $0.14\pm0.08$ & $0.24\pm0.13$ &
$0.09\pm0.41$ & $0.40\pm0.10$ \\
$\gamma=0.2$ & $0.27\pm0.04$ & $0.40\pm0.09$ &
$0.45\pm0.20$ & $0.36\pm0.05$ \\
$\gamma=1$ & $0.31\pm0.10$ & $0.45\pm0.15$ &
$0.40\pm0.27$ & $0.10\pm0.29$ \\
Vision-token & $0.30\pm0.03$ & $0.45\pm0.12$ &
$0.61\pm0.03$ & $0.41\pm0.05$ \\
\bottomrule
\end{tabular}
\end{table}

\paragraph{Recovery is broad but not uniform.}
Moderate coupling improves identity, future displacement, and action transfer,
although action decoding remains below the native vision-token model. Geometry
is non-monotonic under stronger coupling. These results support the narrower
conclusion that a moderate aligned path prevents the collapsed solution and
restores the principal forecast and spatial-transfer properties identified in RQ1.

\subsection{Backward-Only Attention-Routing Intervention}
\label{app:rope-special-credit}

Figure~\ref{fig:rq2-credit}(b) tests whether gradients returned through special-token attention
can train the visual stream while the forward computation remains fixed. This
subsection describes the stable attention route, the backward-only operators,
and a shared-weight control.

\paragraph{Stable attention route.}
The unindexed special-token interface from RQ1 has no consistent
special-token-to-patch correspondence to manipulate. We therefore construct a
diagnostic model with $M=N$ special tokens placed in patch order under 1-D
RoPE. Patch $i$ and special token $i$ have the same relative positional offset
for every $i$. Across all five seeds, each special token attends most strongly
to its corresponding patch in essentially every case. This construction is
used only to isolate the returned attention gradient and is not the RQ1
special-token baseline. The unindexed baseline remains a reference in
Figure~\ref{fig:rq2-credit}(b).

\paragraph{Backward-only intervention.}
All conditions use the same forward activations. In every Transformer block,
each special-token query reads the patch keys and values through an
identity-forward copy of the patch activations. The backward operator either
preserves the native gradient, blocks the gradient before it reaches the patch
activations, or redirects it through a random spatial permutation. A new
permutation is drawn for every sample and training step, while the same
permutation is used across blocks within one forward pass. The shuffled
gradient is therefore consistent through depth but has no stable relation to
the source patch across updates. These operators do not change the forward
value for fixed parameters.

\paragraph{Primary untied setting and shared-weight control.}
The primary intervention in Figure~\ref{fig:rq2-credit}(b) gives the visual and special-token
streams separate Transformer weights and final normalization layers. Returned
attention credit is then the only training signal available to the visual
stream, so blocking it leaves that stream untrained. We additionally evaluate
a shared-weight model in which both streams update the same Transformer.
Blocking still removes the gradient returned to patch activations, but the
visual tokens can learn indirectly through weights updated by the
special-token stream. The shared-weight setting is therefore a control rather
than the primary isolation.

The vision-token reference is retrained with the same 1-D RoPE encoder as the
intervention conditions, while the unindexed special-token reference is the
RQ1 model. These rows provide context and are not intervention conditions.

\begin{table*}[t]
\centering
\small
\caption{Backward-only attention-routing intervention in the RoPE-indexed
$M=N$ special-token model. Values are mean $\pm$ standard deviation over five
training seeds. Gap closed denotes decoded-forecast gap closure. The blocked
shared-weight forecast uses the four runs with available decoded forecasts.}
\label{tab:rope-special-credit}
\setlength{\tabcolsep}{5pt}
\begin{tabular}{@{}lrrrr@{}}
\toprule
& & & \multicolumn{2}{c}{\textbf{Spatial-tail $R^2$}} \\
\cmidrule(lr){4-5}
\textbf{Condition} & \textbf{Gap closed (\%)} & \textbf{Eff. rank} &
\textbf{Position} & \textbf{Action} \\
\midrule
\multicolumn{5}{@{}l}{\emph{Untied special-token stream}} \\
Native credit & $50.8\pm1.6$ & $117.4\pm7.4$ &
$0.395\pm0.060$ & $0.249\pm0.067$ \\
Credit blocked & $42.9\pm3.7$ & $22.9\pm3.8$ &
$0.301\pm0.027$ & $0.342\pm0.105$ \\
Credit shuffled & $14.5\pm15.1$ & $4.1\pm2.7$ &
$0.067\pm0.092$ & $0.133\pm0.206$ \\
\midrule
\multicolumn{5}{@{}l}{\emph{Shared block weights}} \\
Route-enabled, native credit & $43.5\pm0.4$ &
$31.3\pm5.6$ & $0.534\pm0.059$ &
$0.579\pm0.158$ \\
Credit blocked & $31.0\pm3.1$ & $22.8\pm6.4$ &
$0.436\pm0.049$ & $0.580\pm0.077$ \\
Credit shuffled & $10.6\pm11.2$ & $6.5\pm2.8$ &
$0.164\pm0.082$ & $0.163\pm0.116$ \\
\midrule
Vision-token, same RoPE encoder & $34.5\pm5.1$ &
$43.3\pm2.1$ & $0.502\pm0.039$ &
$0.690\pm0.017$ \\
RQ1 special-token, unindexed & $0.3\pm2.9$ &
$6.3\pm5.1$ & $0.138\pm0.134$ &
$-0.031\pm0.401$ \\
\bottomrule
\end{tabular}
\end{table*}

The untied block reports the exact values summarized in Figure~\ref{fig:rq2-credit}(b). The
shared-weight control differs because blocking returned activation credit does
not prevent learning through parameters updated by the special-token stream.
Native and blocked conditions therefore remain closer than in the untied
setting. Shuffling nevertheless reduces both spatial and action readouts,
showing that sensitivity to the gradient destination persists when the two
streams share parameters. The shared-weight results should therefore be read
as a robustness check rather than a second clean isolation.

\begin{figure}[htbp]
    \centering
    \includegraphics[width=\linewidth]{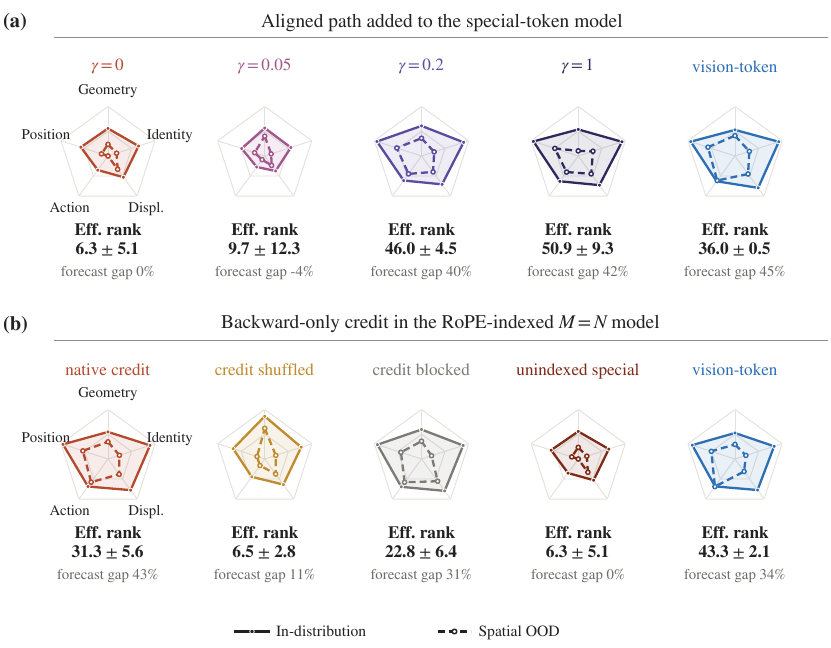}
    \caption{
    \textbf{Complete frozen-probe profiles for the aligned-path and
    backward-only interventions.}
    Each radar reports geometry, identity, future displacement, action chunk,
    and position from the visual-token representation. Solid and dashed
    outlines denote near-center and spatial-tail evaluation, averaged over
    five seeds. Identity uses accuracy and the remaining probes use $R^2$.
    Effective rank and forecast gap are shown alongside each profile.
    \textbf{(a)} Aligned-path strengths and the vision-token reference.
    \textbf{(b)} Native, blocked, and shuffled attention credit in the
    shared-weight control, with the corresponding special-token and
    vision-token references.
    }
    \label{fig:app-intervention-radars}
\end{figure}

\subsection{Degrading the Spatial Resolution of Vision-Token Credit}
\label{app:gradient-routing-intervention}

To complement the attention-routing intervention, we start from the
vision-token model and progressively remove spatial resolution from the
gradient delivered to its encoder. The forward prediction, target, and loss
remain unchanged. If spatially resolved forecast credit is necessary for the
vision-token representation, coarsening this backward update should weaken the
learned spatial and action structure even when its norm is preserved.

Pool-16 averages the patchwise gradients within a $4\times4$ region grid,
Pool-4 within a $2\times2$ grid, and Pool-1 over the full image. Each regional
average is copied back to its patches and rescaled to match the original
sample-wise gradient norm. We additionally include a random permutation of
patch locations and a zero-gradient condition as diagnostic controls.

\begin{table}[htbp]
\centering
\scriptsize
\caption{Effect of spatially coarsening vision-token forecast credit. Values
are mean $\pm$ standard deviation over five seeds. The special-token model is
a reference trained normally with $M=N=64$.}
\label{tab:gradient-intervention-results}
\begin{tabular}{@{}lrrrr@{}}
\toprule
\textbf{Condition} &
\shortstack{\textbf{Gap closed}\\\textbf{(\%)}} &
\textbf{Eff. rank} &
\shortstack{\textbf{Position $R^2$}\\\textbf{(spatial tail)}} &
\shortstack{\textbf{Action $R^2$}\\\textbf{(spatial tail)}} \\
\midrule
Vision-token & $45.0\pm1.2$ & $36.0\pm0.5$ & $0.57\pm0.03$ & $0.61\pm0.03$ \\
Pool-16 & $29.1\pm4.6$ & $37.5\pm6.2$ & $0.50\pm0.10$ & $0.26\pm0.09$ \\
Pool-4 & $1.0\pm1.6$ & $12.6\pm7.9$ & $0.19\pm0.09$ & $0.18\pm0.18$ \\
Pool-1 & $1.2\pm0.7$ & $15.7\pm3.9$ & $0.14\pm0.11$ & $-0.17\pm0.32$ \\
Shuffle & $2.0\pm0.7$ & $14.0\pm1.9$ & $0.17\pm0.13$ & $-0.08\pm0.17$ \\
Block & $20.2\pm4.1$ & $13.5\pm1.1$ & $0.19\pm0.01$ & $0.30\pm0.03$ \\
Special-token & $0.3\pm2.9$ & $6.3\pm5.1$ & $0.14\pm0.13$ & $-0.03\pm0.40$ \\
\bottomrule
\end{tabular}
\end{table}

The pooling sweep shows that the benefit of vision-token forecasting depends
on the spatial resolution of its backward update. Coarsening the update to 16
regions retains part of the forecast and transfer benefit, while four-region
and global updates approach the collapsed special-token profile. Because every
pooled gradient is norm matched, this degradation cannot be explained by
weaker supervision alone. It provides the reverse counterpart to the
aligned-path rescue in Appendix~\ref{app:aligned-path-rescue}.

Shuffling produces a similar failure but is used only as a
location-sensitivity diagnostic because the permuted update is not a descent
direction for the original loss. Blocking is also diagnostic rather than a
lower bound, since the nearly untrained encoder can retain pixel information
through random features. Neither condition is treated as equivalent to the
special-token attention route.

\section{VLA-Scale Simulation Evaluation (RQ3)}
\label{app:vla-scale-evaluation}

At VLA scale, prediction and action are optimized jointly, so the routing
effects isolated in RQ2 matter only if they shape the shared visual
representation used for control. We compare policy variants with a shared
backbone and training protocol under distribution shift. We then examine how
stable spatial addressing and
scene-matched future content affect the delivery of predictive supervision and
the resulting visual representation.

\subsection{VLA Variants and Shared Training Protocol}
\label{app:vla-policy-training-evaluation}

\paragraph{Shared training setup.}
All $\pi_{0.5}$ variants start from the same pretrained checkpoint and use
demonstrations from five LIBERO training collections, including LIBERO-90.
Evaluation follows the four-suite LIBERO and LIBERO-PRO protocol described in
Section~\ref{app:vla-policy-results}. The variants are fine-tuned for
10,000 steps with batch size 32, peak learning rate $1.2\times10^{-4}$, 200
warmup steps, and action horizon 16. The action-only baseline retains the
original flow-matching objective. All forecasting variants use the same
momentum target-encoder construction (EMA SigLIP, $m=0.996$), residual target
$z(o_{t+h})-z(o_t)$, horizon $h=16$, motion-weighted $\ell_2$ cost, auxiliary
weight $\lambda=0.1$, and isolated placement of the prediction tokens.

\paragraph{Prediction interfaces.}
The special-token interface appends $M=16$ special tokens and predicts a
$4\times4$ region-pooled future residual. The vision-token interface instead
predicts a residual for every visual patch directly from the prefix vision
tokens. The shuffled-future control retains this vision-token route while
permuting future targets across the batch, separating spatial attachment from
scene-matched future content. Each interface uses a 2048-wide MLP prediction
head. The token-shared special-token head has the same $8.4$M parameters for
$M=16$ and $M=N=256$, while the vision-token head has $1.1$M parameters.

\paragraph{Granularity-matched special-token control.}
The native $M=16$ special-token interface uses 16 region-level targets, whereas
the vision-token interface predicts all $N=256$ patch targets. To separate
forecast granularity from the prediction route, we train a special-token
control with one token for each visual patch ($M=N=256$). Special token $k$ is
supervised against the residual of patch $k$. Its prediction head, attention
isolation, targets, and optimization settings otherwise match the $M=16$
interface.

\begin{figure}[htbp]
    \centering
    \includegraphics[width=\linewidth]{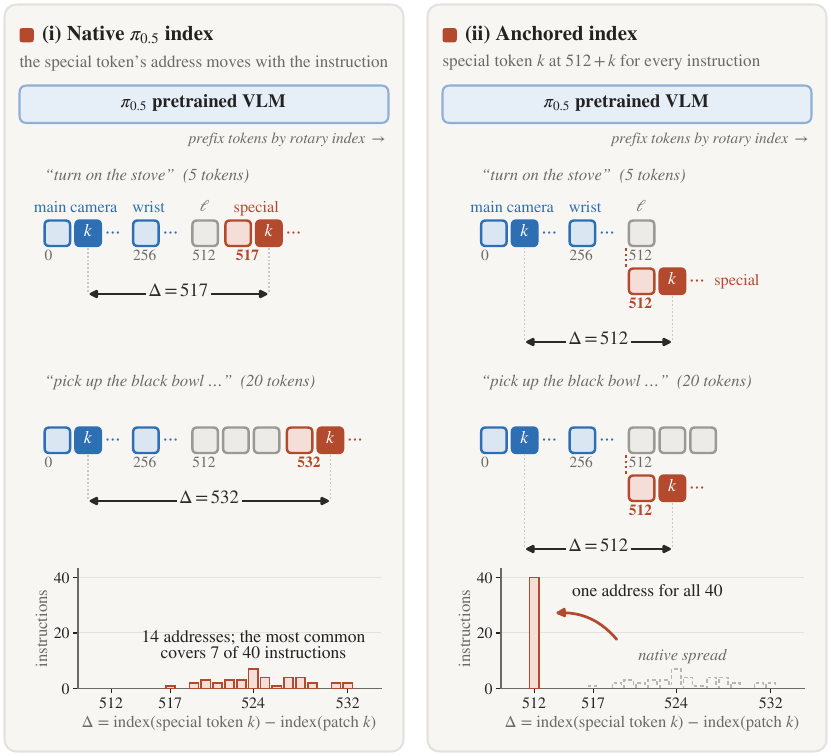}
    \caption{\textbf{Anchored rotary indexing removes prompt-dependent patch
    offsets.} Native cumulative indexing places the special-token stream after
    the instruction, so the offset between corresponding special tokens and
    patches changes with prompt length. Anchoring assigns special token $k$ the
    index $512+k$, preserving an offset of 512 for every prompt. The bottom plot
    shows the offsets induced by the training instructions, with the native
    distribution repeated for comparison.}
    \label{fig:anchored-index}
\end{figure}

\paragraph{Anchored-index special-token control.}
Matching the number of tokens does not by itself provide a stable positional
correspondence in $\pi_{0.5}$. The model assigns rotary position indices
cumulatively over valid prefix tokens, and the special-token stream follows
the variable-length instruction. The relative index between special token $k$
and main-camera patch $k$ is therefore $512+n_{\text{lang}}$. It ranges from
517 to 532 across the 40 training instructions and reaches 533 for the
rewritten LIBERO-PRO instructions. Special token $k$ is supervised against
patch $k$, but its rotary offset from that patch changes with the prompt
(Figure~\ref{fig:anchored-index}).

The anchored control changes only this positional addressing. We assign
special token $k$ the index $512+k$, exactly 512 positions after main-camera
patch $k$ for every instruction. Image and language indices remain unchanged,
and the action-expert suffix is shifted to follow the resulting prefix. The
three training seeds otherwise repeat the $M=N=256$ control, and evaluation
uses the same indexing rule. We verified that training and evaluation construct
identical indices. Although the anchored indices overlap with instruction-token
indices, language-perturbation performance remains within the seed range of the
variable-index control.

\subsection{Evaluation Protocol and Additional Results}
\label{app:vla-policy-results}

We evaluate the standard condition on four LIBERO suites and follow the
LIBERO-PRO protocol~\citep{Zhou2026-gm} for five perturbation families that
modify language, object position, object attributes, task specification, and
environment layout. The resulting 24 suite-condition cells use ten trials per
task under a fixed evaluation seed. Retention is computed within each suite as
perturbed success divided by unperturbed success and then averaged across
suites. Prefix tokens use the same attention isolation at training and
evaluation. Reported uncertainty is one standard deviation over three training
seeds.

Table~\ref{tab:rq3-main} reports the aggregate $\pi_{0.5}$ results. Across matched training
seeds, vision-token forecasting consistently improves mean LIBERO-PRO success
over both the baseline and the $M=16$ special-token interface. Matching the
number of special tokens to visual patches does not produce a consistent gain,
whereas anchoring their spatial indices improves both object-position shift
and mean LIBERO-PRO success in all three seeds.
Table~\ref{tab:rq3-per-suite} provides the suite-level breakdown behind these
aggregate results.

\begin{table}[t]
\centering
\caption{\textbf{Per-suite $\pi_{0.5}$ simulation results.} Success rate (\%)
for every suite and perturbation, averaged over three training seeds. Averaging
each column over the four suites recovers the corresponding result in Table~\ref{tab:rq3-main}.}
\label{tab:rq3-per-suite}
\small
\setlength{\tabcolsep}{5pt}
\begin{tabular}{lcccccc}
\toprule
& LIBERO
& \multicolumn{5}{c}{LIBERO-PRO} \\
\cmidrule(lr){2-2} \cmidrule(lr){3-7}
\textbf{Variant / suite}
& none
& lang.
& pos.
& obj.
& task
& env. \\
\midrule
\multicolumn{7}{l}{\textit{Baseline (no forecast)}} \\
\quad Spatial & $95.7$ & $94.0$ & $70.0$ & $69.3$ & $56.7$ & $51.0$ \\
\quad Object & $99.0$ & $97.7$ & $60.7$ & $87.0$ & $43.7$ & $66.0$ \\
\quad Goal & $93.7$ & $91.3$ & $44.7$ & $85.8$ & $45.7$ & $58.3$ \\
\quad LIBERO-10 & $81.0$ & $61.0$ & $20.0$ & $47.0$ & $17.0$ & $22.7$ \\
\addlinespace[2pt]
\multicolumn{7}{l}{\textit{Special-token ($M{=}16$)}} \\
\quad Spatial & $93.0$ & $92.0$ & $65.7$ & $74.3$ & $60.0$ & $47.7$ \\
\quad Object & $98.0$ & $96.3$ & $54.0$ & $89.0$ & $43.0$ & $73.0$ \\
\quad Goal & $96.0$ & $90.3$ & $40.0$ & $95.0$ & $46.0$ & $61.3$ \\
\quad LIBERO-10 & $81.0$ & $57.3$ & $24.0$ & $58.1$ & $23.7$ & $24.7$ \\
\addlinespace[2pt]
\multicolumn{7}{l}{\textit{Special-token ($M{=}N{=}256$)}} \\
\quad Spatial & $93.0$ & $92.3$ & $66.3$ & $76.3$ & $58.3$ & $52.3$ \\
\quad Object & $99.3$ & $98.0$ & $49.7$ & $85.7$ & $38.7$ & $71.3$ \\
\quad Goal & $90.7$ & $86.7$ & $42.0$ & $92.5$ & $48.3$ & $58.3$ \\
\quad LIBERO-10 & $79.7$ & $65.7$ & $23.0$ & $51.9$ & $20.4$ & $26.7$ \\
\addlinespace[2pt]
\multicolumn{7}{l}{\textit{Special-token ($M{=}N{=}256$, anchored)}} \\
\quad Spatial & $97.3$ & $91.0$ & $74.0$ & $84.3$ & $53.3$ & $53.0$ \\
\quad Object & $98.7$ & $98.0$ & $69.7$ & $94.0$ & $48.7$ & $67.7$ \\
\quad Goal & $92.3$ & $85.7$ & $48.3$ & $87.5$ & $52.0$ & $57.0$ \\
\quad LIBERO-10 & $82.0$ & $68.3$ & $26.0$ & $55.2$ & $23.3$ & $32.0$ \\
\addlinespace[2pt]
\multicolumn{7}{l}{\textit{Vision-token}} \\
\quad Spatial & $96.7$ & $96.0$ & $77.3$ & $77.3$ & $56.2$ & $57.0$ \\
\quad Object & $98.7$ & $97.0$ & $82.7$ & $82.7$ & $53.7$ & $80.0$ \\
\quad Goal & $95.0$ & $89.0$ & $51.3$ & $94.2$ & $52.7$ & $60.0$ \\
\quad LIBERO-10 & $88.3$ & $71.3$ & $32.3$ & $56.3$ & $21.5$ & $38.0$ \\
\addlinespace[2pt]
\multicolumn{7}{l}{\textit{Vision-token (shuffled future)}} \\
\quad Spatial & $94.7$ & $92.0$ & $63.7$ & $77.7$ & $61.2$ & $47.0$ \\
\quad Object & $98.3$ & $98.7$ & $48.7$ & $88.7$ & $34.0$ & $69.7$ \\
\quad Goal & $94.7$ & $91.7$ & $38.3$ & $88.3$ & $43.0$ & $60.7$ \\
\quad LIBERO-10 & $81.0$ & $57.7$ & $20.7$ & $51.9$ & $17.0$ & $33.3$ \\
\bottomrule
\end{tabular}
\end{table}

The position result is consistent across suites. Vision-token forecasting
exceeds the baseline in all four, and the anchored control exceeds its
variable-index counterpart in all four. Individual suite cells are otherwise
reported as a complete breakdown rather than as separate statistical claims.

\paragraph{SmolVLA replication.}
We repeat the native interface comparison on
SmolVLA~\citep{Shukor2025-smolvla}, a 450M-parameter policy. The variants are
fine-tuned for 30,000 steps with the same residual targets, motion weighting,
auxiliary weight, and evaluation protocol. Table~\ref{tab:smolvla-results}
reports the complete perturbation breakdown.

\begin{table}[t]
\centering
\caption{\textbf{SmolVLA simulation results.} Values are mean $\pm$ standard
deviation over three training seeds. Retention is perturbed success divided by
unperturbed success. Best values are bold.}
\label{tab:smolvla-results}
\small
\setlength{\tabcolsep}{4pt}
\begin{tabular}{lcccccccc}
\toprule
& LIBERO
& \multicolumn{6}{c}{LIBERO-PRO}
& \\
\cmidrule(lr){2-2} \cmidrule(lr){3-8}
\textbf{Variant}
& \textbf{mean}
& lang.
& pos.
& obj.
& task
& env.
& \textbf{mean}
& \textbf{Ret.} \\
\midrule
Baseline
& $88.2{\scriptstyle\,\pm 1.1}$
& $22.1$ & $0.0$ & $47.3$ & $13.9$ & $15.1$
& $19.7{\scriptstyle\,\pm 1.5}$ & $0.221$ \\
Special-token
& $90.7{\scriptstyle\,\pm 2.5}$
& $22.8$ & $0.0$ & $45.2$ & $\mathbf{14.8}$ & $\mathbf{17.4}$
& $20.0{\scriptstyle\,\pm 0.6}$ & $0.220$ \\
Vision-token
& $\mathbf{92.8}{\scriptstyle\,\pm 0.7}$
& $\mathbf{32.1}$ & $\mathbf{0.1}$ & $\mathbf{52.4}$ & $13.7$ & $15.2$
& $\mathbf{22.7}{\scriptstyle\,\pm 0.9}$ & $\mathbf{0.243}$ \\
\bottomrule
\end{tabular}
\end{table}

Vision-token forecasting exceeds special-token forecasting on all three
training seeds. Because performance under object-position perturbation is near
zero for every variant, this experiment supports the interface ordering but
does not provide an additional test of the spatial mechanism.

\subsection{Anchoring Stabilizes Spatial Credit Delivery}
\label{app:vla-m256-delivery}

The anchored control is designed to make each special token consistently
address the corresponding visual patch. We test whether this change also
affects how the forecast loss reaches the visual stream. For a target
associated with one of the 16 image regions, we measure the fraction of its
gradient norm assigned to vision tokens in the same region at the visual
prefix and intermediate Transformer layers. A spatially unstructured route has
an expected share of $1/16$.

Measurements use 60 unperturbed demonstration frames from each of the four
LIBERO suites. Both $M=N=256$ variants are averaged over three training seeds.
The $M=16$ and vision-token references use one checkpoint each.
Table~\ref{tab:vla-m256-delivery} reports how strongly each interface directs
forecast gradients to the corresponding image region across depth. The two
$M=N=256$ rows isolate the effect of anchoring while holding token count and
patchwise targets fixed.

\begin{table}[t]
\centering
\small
\caption{\textbf{Spatial delivery of forecast supervision.} Same-region share
of the forecast-loss gradient across the $\pi_{0.5}$ visual stream. Chance is
$1/16$. The $M=N=256$ rows average three training seeds. Gradients use the
final online SigLIP features as targets because the training-time EMA target
encoder was not retained.}
\label{tab:vla-m256-delivery}
\setlength{\tabcolsep}{6pt}
\begin{tabular}{@{}lccccc@{}}
\toprule
\textbf{Interface} & \textbf{Prefix} & \textbf{L6} & \textbf{L11} &
\textbf{L14} & \textbf{L17} \\
\midrule
Special-token, $M=16$ & $0.13$ & $0.10$ & $0.12$ & $0.12$ & $0.08$ \\
Special-token, $M=N=256$ & $0.31$ & $0.43$ & $0.55$ & $0.61$ & $0.29$ \\
Special-token, $M=N=256$, anchored & $0.55$ & $0.67$ & $0.77$ & $0.80$ & $0.82$ \\
Vision-token & $0.70$ & $0.80$ & $0.88$ & $0.91$ & $0.96$ \\
\bottomrule
\end{tabular}
\end{table}

The key comparison is between the two $M=N=256$ rows. With variable
indices, spatial alignment emerges in the middle of the Transformer but is
weaker at the visual prefix and falls again in the final measured layer.
Anchoring produces a consistently aligned profile from the prefix through the
later layers. The $M=16$ and vision-token rows provide lower and upper
references for this change.

Patchwise targets alone therefore do not ensure a stable spatial route to the
visual stream. Anchoring makes this delivery persistent across depth.
Section~\ref{app:vla-layer-probes} next examines whether that change is
reflected in the information encoded by the visual representation.

\subsection{Information About Future Change in the VLA Visual Stream}
\label{app:vla-layer-probes}

Having established that anchoring stabilizes spatial credit delivery, we next
ask what information becomes more accessible in the visual stream. Future
object displacement is the primary target because it measures sensitivity to
upcoming scene change. Object position and object-to-gripper geometry serve as
controls for general spatial readout quality.

\paragraph{Probe setup.}
The primary comparison evaluates all 12 checkpoints from the baseline,
$M=16$ special-token, vision-token, and shuffled-future variants on a shared
frame bank. The bank covers \textsc{LIBERO-Spatial},
\textsc{LIBERO-Object}, \textsc{LIBERO-Goal}, and \textsc{LIBERO-10} under
unperturbed, object-position, and environment conditions. At most 3,000 common
frames are sampled for each probe so that every checkpoint is evaluated on the
same observations.

We read the 256 main-view image tokens at the language-model input and at
layers $2,4,\ldots,18$. Grid readouts average-pool the $16\times16$ map to
$4\times4$ before fitting ridge regression. Targets include object position,
object-to-gripper geometry, object displacement after 16 steps, and
end-effector displacement after 16 steps. Probes are trained on four of five
task groups and evaluated on the held-out group. \textsc{LIBERO-10} is excluded
from these held-out-task scores because changing its task also changes the
scene and object set, producing negative $R^2$ across targets in the baseline.

\paragraph{Target content determines the readout of future change.}
Table~\ref{tab:vla-layer-probes} reports final-layer probes on held-out tasks
under object-position shift.

\begin{table}[t]
\centering
\small
\caption{\textbf{Final-layer visual-token probes under object-position shift.}
Values are probe $R^2$, reported as mean $\pm$ standard deviation over three
training seeds and three LIBERO suites.}
\label{tab:vla-layer-probes}
\begin{tabular}{@{}lrrrr@{}}
\toprule
\textbf{Target} & \textbf{Baseline} & \textbf{Special-token} &
\textbf{Vision-token} & \textbf{Shuffled future} \\
\midrule
Object displacement & $0.360\pm0.024$ & $0.428\pm0.030$ &
$\mathbf{0.504\pm0.003}$ & $0.419\pm0.026$ \\
EEF displacement & $0.738\pm0.020$ & $0.759\pm0.007$ &
$\mathbf{0.793\pm0.010}$ & $0.765\pm0.004$ \\
Object position & $0.413\pm0.049$ & $0.450\pm0.048$ &
$\mathbf{0.545\pm0.021}$ & $0.539\pm0.020$ \\
Object-to-gripper geometry & $0.508\pm0.029$ & $0.543\pm0.045$ &
$\mathbf{0.621\pm0.046}$ & $0.613\pm0.013$ \\
\bottomrule
\end{tabular}
\end{table}

Vision-token forecasting provides the strongest readout of future object and
end-effector displacement. Vision-token and shuffled-future forecasting are
nearly tied on current object position and object-to-gripper geometry, despite
their separation on future displacement. Their shared attachment route can
therefore strengthen current spatial information, while scene-matched future
content is required for the strongest dynamics representation.

\paragraph{Anchoring improves the readout of future change.}
To isolate the effect of stable spatial addressing, we apply the same probes to
the anchored and variable-index $M=N=256$ controls on
\textsc{LIBERO-Spatial} and \textsc{LIBERO-Object}.
Table~\ref{tab:vla-anchored-representation} summarizes the matched
three-seed comparison.

\begin{table}[t]
\centering
\small
\caption{\textbf{Representation probes for the $M=N=256$ controls.} Values are
mean $\pm$ standard deviation over three matched seeds. Future displacement is
reported across layers, with current object position at layer 12 as a control.}
\label{tab:vla-anchored-representation}
\setlength{\tabcolsep}{4pt}
\begin{tabular}{@{}lrrrr@{}}
\toprule
& \multicolumn{3}{c}{\textbf{Future displacement $R^2$}} &
\textbf{Position $R^2$} \\
\cmidrule(lr){2-4} \cmidrule(lr){5-5}
\textbf{Variant} & \textbf{L10} & \textbf{L12} & \textbf{L14} &
\textbf{L12} \\
\midrule
Variable index
& $0.353\pm0.040$ & $0.410\pm0.052$ & $0.430\pm0.035$
& $0.473\pm0.101$ \\
Anchored index
& $\mathbf{0.460\pm0.040}$ & $\mathbf{0.467\pm0.025}$
& $\mathbf{0.467\pm0.025}$ & $0.357\pm0.025$ \\
\bottomrule
\end{tabular}
\end{table}

Anchoring improves future-displacement decoding in every matched seed at layer
10, and its mean advantage remains through layers 12 and 14. Current object
position readout does not improve. The change is therefore specific to future
dynamics rather than a general increase in linear readout quality.
Across the 12 checkpoints in the primary four-variant comparison, layer-12
displacement decoding is strongly associated with object-position-shift
success ($r=0.82$). This association is consistent with, but does not
establish, a role for visual features that retain information about future
scene change in robustness.

\section{Real-Robot Evaluation (RQ3)}
\label{app:real-robot-evaluation}

Physical evaluation tests whether the interface ordering observed in simulation
persists under changes in visual evidence and object-container layout. We
compare five $\pi_{0.5}$ policies on matched real-robot blocks and analyze their
representations on held-out episodes. The study tests transfer to physical
control rather than independently isolating the prediction interface.

\subsection{Policies and Training Data}
\label{app:real-robot-setup}
\label{app:real-robot-data}

All policies start from the same pretrained checkpoint. The action-only
baseline has no forecast objective. The native special-token policy predicts
future features from $M=16$ special tokens. Two granularity-matched controls
use $M=N=256$ special tokens with either the native or anchored indices from
Appendix~\ref{app:vla-policy-training-evaluation}. The vision-token policy predicts directly from the main-camera
patch tokens. Forecasting variants predict main-camera features $0.8$\,s
ahead. The wrist camera remains part of the policy input but receives no
prediction target. All policies use the same data split, random seed,
optimization schedule, action horizon, and 8,000-step evaluation checkpoint.

The dataset contains 442 teleoperated episodes collected at 30\,Hz on a
single-arm Trossen AI Solo robot. A fixed main camera and a wrist camera provide
the policy inputs. The demonstrations cover three tasks: picking a named object
from clutter, picking two named objects in the instructed order, and opening a
pot to place an object inside.

\begin{table}[t]
\centering
\small
\caption{\textbf{Real-robot training data.} The held-out episodes are excluded
from policy training and used for every representation probe.}
\label{tab:real-robot-dataset}
\begin{tabular}{@{}lrrr@{}}
\toprule
\textbf{Task} & \textbf{Training} & \textbf{Held out} &
\textbf{Median length} \\
\midrule
Pick from clutter & 219 & 24 & $7.6$\,s \\
Pick two in order & 89 & 10 & $13.4$\,s \\
Open pot and place & 90 & 10 & $15.7$\,s \\
\midrule
Total & 398 & 44 & \\
\bottomrule
\end{tabular}
\end{table}

\begin{figure}[t]
    \centering
    \includegraphics[width=\linewidth]{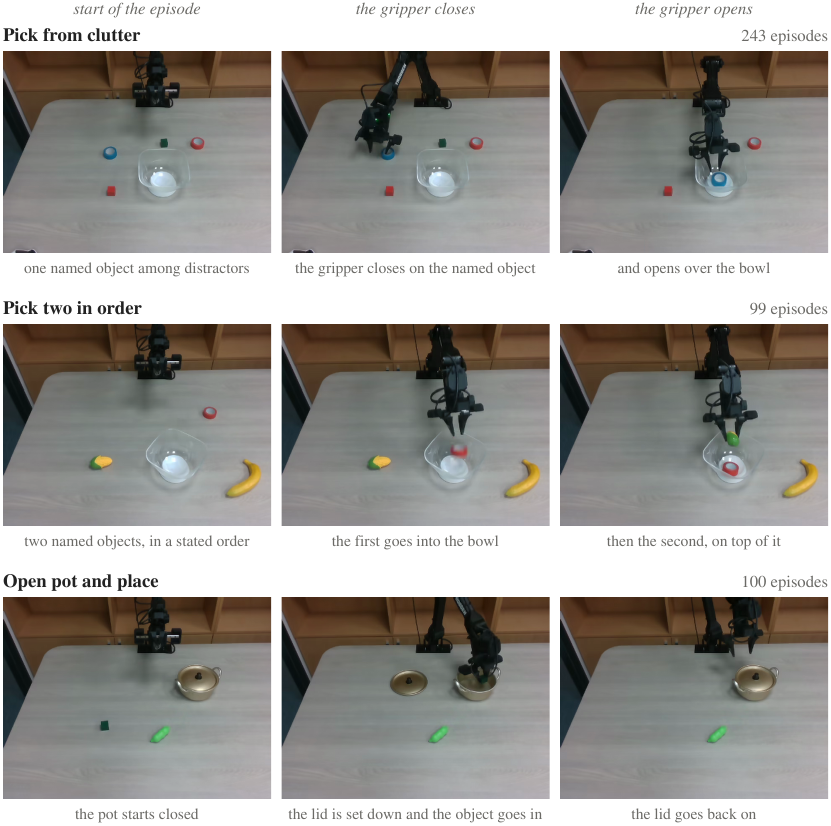}
    \caption{\textbf{The three collection tasks.} Each row shows the start
    frame and the two gripper events from one demonstration. The pot task adds
    a small reflective lid knob and a longer manipulation sequence to the
    object-placement behavior shared by the tasks.}
    \label{fig:real-robot-tasks}
\end{figure}

\begin{figure}[t]
    \centering
    \includegraphics[width=\linewidth]{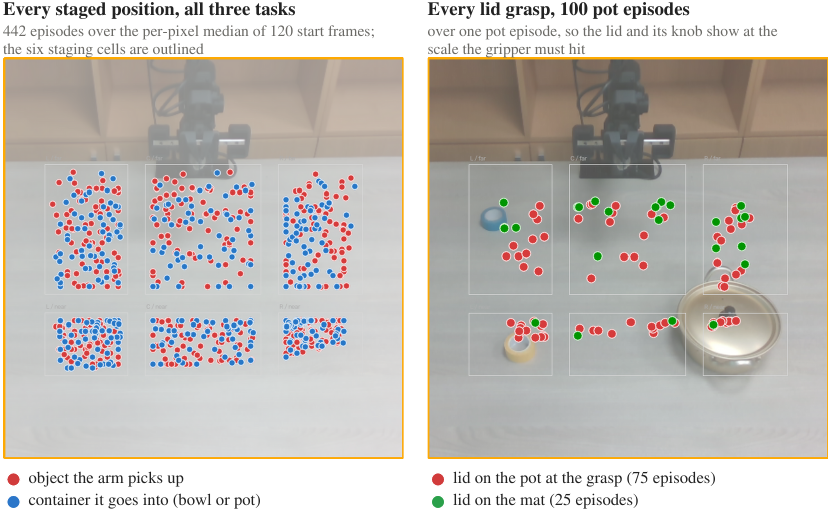}
    \caption{\textbf{Spatial coverage of the training data.} Left: staged
    object and container positions across the 442 demonstrations. Right:
    positions at which the robot grasps the pot lid, shown at the scale of the
    lid and its knob.}
    \label{fig:real-robot-placements}
\end{figure}

Generated staging sheets assign targets, containers, and distractors across six
cells on the workbench. This balances object and container positions while
covering transfers between the left, center, and right regions. The pot
position is staged, but its orientation is not, so the lid knob varies across
episodes. Figure~\ref{fig:real-robot-placements} summarizes the resulting
spatial coverage.

We reserve 44 episodes at the episode level for all representation probes. The split is stratified by task, object identity, spatial
layout, episode length, and recording order. This prevents temporally adjacent
frames from the same trajectory from appearing in both training and
evaluation.

\subsection{Blind Evaluation Protocol and Conditions}
\label{app:real-robot-protocol}
\label{app:real-robot-conditions}

Each evaluation block fixes one instruction and staged layout. Every policy
runs three trials with matched flow-matching noise, and anonymous policy order
is rotated within the block. The operator sees only an anonymous slot label.
Start layouts are restored from reference photographs so that comparisons
remain paired by block.

The three primary policies were evaluated in the original blinded sittings.
The two $M=N=256$ controls were subsequently evaluated on the same saved
layouts, and all five policies were run together on three additional pot
blocks. The final comparison contains 28 matched blocks and 420 trials. Five
additional blocks with the lid starting on the mat were excluded before
scoring because their layouts could not be reproduced reliably.

The primary outcome is task success, averaged within each block. Intermediate
task stages are retained to localize failures. Confidence intervals resample
blocks, and pairwise comparisons use a two-sided exact paired sign test over
blocks with different policy scores.

\begin{table}[t]
\centering
\small
\caption{\textbf{Real-robot evaluation conditions.} Each block is one matched
instruction and layout evaluated three times by every policy.}
\label{tab:real-robot-blocks}
\begin{tabular}{@{}llrl@{}}
\toprule
\textbf{Condition} & \textbf{Task} & \textbf{Blocks} &
\textbf{Changed factor} \\
\midrule
In distribution & bowl tasks & 7 & none \\
Blurred camera & bowl tasks & 10 & main-camera image \\
Unseen pot layout & pot task & 11 & layout and lid grasp \\
\bottomrule
\end{tabular}
\end{table}

\begin{figure}[t]
    \centering
    \includegraphics[width=\linewidth]{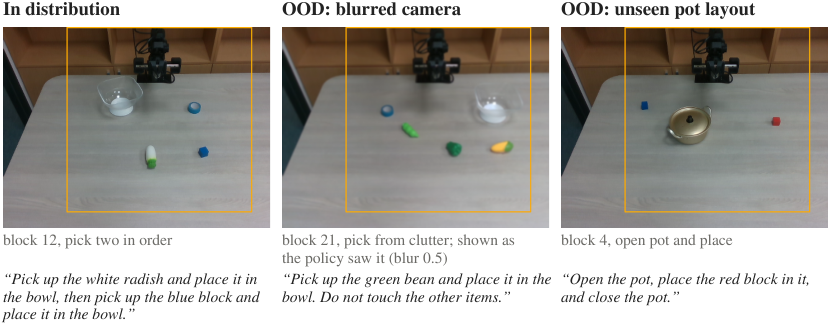}
    \caption{\textbf{Real-robot evaluation conditions.} One matched block from
    the in-distribution, blurred-camera, and unseen-pot conditions. The orange
    square marks the main-camera crop supplied to the policy.}
    \label{fig:real-robot-conditions}
\end{figure}

Figure~\ref{fig:real-robot-conditions} shows the three conditions used in the
pooled analysis. The in-distribution condition follows the same staging process
as the training data. The blurred-camera condition filters only the main-camera
input while leaving the wrist image and physical scene unchanged. The unseen
pot condition requires the policy to localize and grasp the small lid knob
before placing the target object.

We quantify the pot shift from recorded start frames using the detected target
and container positions. Each evaluation layout is compared with its nearest
training layout. Applying the same detector to both sets controls for
localization error.

\begin{table}[t]
\centering
\small
\caption{\textbf{Distance of evaluation layouts from training.} Distances are
measured in the main-camera image and converted to millimeters.}
\label{tab:real-robot-layout-distance}
\begin{tabular}{@{}lrrr@{}}
\toprule
\textbf{Task} & \textbf{Eval. rounds} & \textbf{Eval. median} &
\textbf{Training median} \\
\midrule
Pot & 33 & $51$\,mm & $27$\,mm \\
Bowl pick & 19 & $32$\,mm & $37$\,mm \\
\bottomrule
\end{tabular}
\end{table}

The pot layouts lie roughly twice as far from training as training layouts lie
from one another, while the bowl layouts do not. The pot condition is therefore
a composite physical shift involving object-container layout, the uncommon lid
grasp, and a longer manipulation sequence. We do not interpret it as an
isolated position intervention.

\subsection{Full Policy Results and Failure Diagnostics}
\label{app:real-robot-results}

\begin{table}[t]
\centering
\footnotesize
\setlength{\tabcolsep}{5pt}
\caption{\textbf{Blind real-robot success.} Each entry gives success
percentage with a 95\% interval from resampling staged blocks. The number of
blocks appears in parentheses, and each policy runs three trials per block.
OOD pools the blurred-camera and unseen-pot conditions. Bold marks the highest
point estimate in each column.}
\label{tab:real-robot-full}
\begin{tabular}{@{}lcccc@{}}
\toprule
\textbf{Policy} & \textbf{ID (7)} & \textbf{Blur (10)} &
\textbf{Pot (11)} & \textbf{OOD (21)} \\
\midrule
Baseline & $71\,[57,86]$ & $53\,[33,73]$ & $39\,[21,58]$ &
$46\,[32,60]$ \\
Special, $M=16$ & $86\,[71,95]$ & $73\,[50,90]$ & $33\,[15,52]$ &
$52\,[37,68]$ \\
Special, $M=N=256$ & $76\,[57,90]$ & $\mathbf{83}\,[67,100]$ &
$30\,[9,52]$ & $56\,[38,73]$ \\
Special, $M=N=256$, anchored & $81\,[62,95]$ & $77\,[50,100]$ &
$42\,[27,61]$ & $59\,[43,75]$ \\
Vision-token & $\mathbf{95}\,[86,100]$ & $80\,[63,97]$ &
$\mathbf{64}\,[42,82]$ & $\mathbf{71}\,[57,84]$ \\
\bottomrule
\end{tabular}
\end{table}

Table~\ref{tab:real-robot-full} reports the complete block-level results.
Across the 21 shifted blocks, vision-token forecasting wins against the
baseline on 11 discordant blocks without a loss ($p=0.001$) and against the
$M=16$ special-token policy on 11 blocks with two losses ($p=0.022$). Its
comparisons with the two $M=N=256$ controls are not resolved. The pooled paired
analysis therefore resolves the two primary comparisons but not the comparisons
among all forecasting variants.

The two $M=N=256$ controls remain competitive under camera blur but return to
the baseline range on the unseen pot layouts, with little difference between
native and anchored indices. This condition dependence explains why their
pooled performance lies between the baseline and vision-token policy without
isolating an anchored-index effect on the robot.

The stage annotations localize the pot failures. On the eight original pot
blocks, vision-token forecasting removes the lid in 22 of 24 trials, compared
with 17 for the baseline and 15 for the $M=16$ special-token policy. The
performance gap therefore appears during the unfamiliar lid interaction,
before object placement is completed.

\subsection{Held-Out Representation Diagnostics}
\label{app:real-robot-probes}

We freeze each policy and fit ridge readouts to its average-pooled main-camera
patch tokens. All readouts use the same 44 episodes excluded from policy
training, with cross-validation grouped by episode. We decode gripper width
and joint motion $0.8$\,s into the future, while present position and
joint-state readouts control for a general improvement in linear-probe
performance.
Table~\ref{tab:real-robot-probes} reports the 8,000-step checkpoint. The three
primary policies follow the same ordering at 4,000 steps.

\begin{table}[t]
\centering
\footnotesize
\setlength{\tabcolsep}{4pt}
\caption{\textbf{Information retained in the real-robot visual stream.}
Ridge readouts use main-camera tokens and episode-grouped cross-validation on
the held-out episodes. Special 16 and Special 256 use native special-token
indices, while Anchored 256 uses anchored indices. Higher is better for $R^2$,
and lower is better for position error. Bold marks the best point estimate in
each row. Uncertainty statements in the text use 95\% bootstrap intervals over
episodes.}
\label{tab:real-robot-probes}
\begin{tabular}{@{}lrrrrr@{}}
\toprule
\textbf{Readout} & \textbf{Baseline} & \textbf{Special 16} &
\textbf{Special 256} & \textbf{Anchored 256} & \textbf{Vision} \\
\midrule
\multicolumn{6}{@{}l}{\emph{Future, $0.8$\,s ahead ($R^2$)}} \\
Gripper width & 0.435 & 0.606 & 0.561 & 0.566 & \textbf{0.688} \\
Joint motion & 0.647 & 0.735 & 0.701 & 0.713 & \textbf{0.749} \\
\midrule
\multicolumn{6}{@{}l}{\emph{Present state}} \\
Container position (mm error) & 57.0 & 53.5 & 49.5 & \textbf{48.7} & 49.2 \\
Object position (mm error) & 58.0 & 56.1 & 60.5 & 55.2 & \textbf{54.1} \\
Joint angles ($R^2$) & 0.906 & \textbf{0.909} & 0.883 & 0.889 & 0.902 \\
\bottomrule
\end{tabular}
\end{table}

The contrast is clearest for information about future behavior. Vision-token
forecasting is highest on both future readouts and is separated from the
baseline on each. Its advantage over the $M=16$ special-token policy is also
resolved for future gripper width, but not for future joint motion. The
$M=N=256$ controls instead reach the vision-token level on current container
position while remaining weaker on the future readouts. Object-position
differences are unresolved, and no forecasting policy improves present
joint-angle decoding. The result is therefore not a uniform improvement in
linear-probe performance.

The horizon and token-stream controls further localize the difference. The
vision-token advantage in future gripper width is largest at the trained
$0.8$\,s horizon and falls by $2.4$\,s, where it is no longer resolved. At
the trained horizon, the vision-minus-baseline gap is $0.252$ in the
main-camera tokens but only $0.021$ in the language tokens, where the interval
includes zero. The additional future information is thus concentrated in the
visual stream to which the forecasting objective is attached.

\section{Formal Analysis of Prediction Credit Routes}
\label{app:derivation-gradients}

This section formalizes two distinctions used in the experiments. The first is
the explicit same-position path provided by vision-token readout. The second is
the stable relative address provided by anchored rotary indices. Throughout
this section, prediction credit denotes the gradient signal induced by the
future-prediction loss at a visual-token activation. A credit route is the
Jacobian path through which that signal reaches the visual stream. We isolate
the future-prediction objective because gradients from the action objective are
additive and do not alter this decomposition.

For clarity, we first write one prediction carrier for each spatial target.
This matches the controlled comparison and the $M=N$ VLA controls. The $M=16$
region-pooled interface uses a different target granularity, but its
special-token carriers likewise lack a visual residual identity path.

\subsection{General Token-Level Decomposition}

Let $V=[v_1,\ldots,v_N]^\top$ denote the current visual tokens and
$Z=[z_1,\ldots,z_N]^\top$ the representations carrying predictions. A
token-wise head produces $\hat r_i=H_\psi(z_i)$ with per-target loss
$\ell_i=\mathcal D(\hat r_i,r_i)$ and
$\mathcal L_{\mathrm{fut}}=N^{-1}\sum_i\ell_i$. The teacher target $r_i$
is stop-gradient. Define
\begin{equation}
u_i := \frac{\partial\ell_i}{\partial z_i},
\qquad
B_{ij} := \frac{\partial z_i}{\partial v_j}.
\label{eq:app-interface-jacobian}
\end{equation}
The contribution of prediction $i$ to visual token $j$ and the total auxiliary
gradient at that token are
\begin{equation}
\frac{\partial\ell_i}{\partial v_j}=B_{ij}^{\top}u_i,
\qquad
g_j:=\frac{\partial\mathcal{L}_{\mathrm{fut}}}{\partial v_j}
=\frac{1}{N}\sum_{i=1}^{N}B_{ij}^{\top}u_i.
\label{eq:app-token-gradient}
\end{equation}
The upstream error $u_i$ captures the target and prediction head, while
$B_{ij}$ captures how the interface delivers that error to the visual
sequence.

\subsection{Direct and Attention-Mediated Credit Routes}

Let $C$ denote the remaining context tokens. For a residual vision-token
stream, write
$z_i^{\mathrm{vis}}=v_i+F_{\phi,i}(V,C)$. Its block Jacobian is
\begin{equation}
B_{ij}^{\mathrm{vis}}=\delta_{ij}I+J_{ij}^{F},
\qquad
J_{ij}^{F}:=\frac{\partial F_{\phi,i}(V,C)}{\partial v_j}.
\end{equation}
The identity term provides an explicit same-position component in addition to
the contextual routes.

For the special-token stream $Q=[q_1,\ldots,q_N]^\top$, write
$z_i^{\mathrm{sp}}=q_i+A_{\phi,i}(Q,V,C)$. Since $q_i$ is independent of the
visual sequence,
\begin{equation}
B_{ij}^{\mathrm{sp}}=J_{ij}^{A},
\qquad
J_{ij}^{A}:=\frac{\partial A_{\phi,i}(Q,V,C)}{\partial v_j}.
\end{equation}
The special-token residual provides a direct gradient to $q_i$ but not to a
visual token. In a Transformer, every term in $J_{ij}^{A}$ that connects a
special-token output to a visual-token activation contains at least one
cross-token attention operation. Combining both cases gives
\begin{equation}
\frac{\partial\ell_i}{\partial v_j}=
\begin{cases}
\delta_{ij}u_i+(J_{ij}^{F})^\top u_i,
& \text{vision-token interface},\\[3pt]
(J_{ij}^{A})^\top u_i,
& \text{special-token interface}.
\end{cases}
\label{eq:app-interface-gradient-comparison}
\end{equation}
This comparison does not imply that special tokens provide no visual gradient.
Their gradient reaches the visual stream through attention-mediated token
mixing rather than through an explicit same-position identity term.

\subsection{What RoPE Anchoring Provides}

The decomposition above distinguishes a direct residual route from a route
mediated by attention. We next examine how rotary positional embeddings
structure the latter. Consider one attention head whose query is the
special-token state $x_i^{\mathrm{sp}}$ and whose key is visual token $v_j$.
Define $\tilde q_i=W_Qx_i^{\mathrm{sp}}$ and
$\tilde k_j=W_Kv_j$. Before the softmax, the attention logit is
\begin{equation}
e_{ij}
=
\frac{1}{\sqrt{d_h}}
\left(R(p_i^{\mathrm{sp}})\tilde q_i\right)^\top
\left(R(p_j^{\mathrm{vis}})\tilde k_j\right)
=
\frac{1}{\sqrt{d_h}}
\tilde q_i^\top
R(p_j^{\mathrm{vis}}-p_i^{\mathrm{sp}})
\tilde k_j,
\end{equation}
where $R(p)$ is the rotary transformation and
$R(p)^\top R(p')=R(p'-p)$. RoPE therefore parameterizes the positional
part of this interaction through the relative offset between the two tokens.

Let visual token $v_j$ have rotary index $p_j^{\mathrm{vis}}=j$. In the
native downstream interface, the special tokens follow both visual streams
and the instruction. Their indices are
\begin{equation}
p_{i,\mathrm{native}}^{\mathrm{sp}}
=
D+n_{\mathrm{lang}}+i,
\end{equation}
where $D=512$ is the number of preceding visual positions in our
$\pi_{0.5}$ prefix and $n_{\mathrm{lang}}$ depends on the instruction. Define
the relative offset used by the attention logit as
\begin{equation}
\Delta_{ij}^{\mathrm{native}}
=
p_j^{\mathrm{vis}}-p_{i,\mathrm{native}}^{\mathrm{sp}}
=
j-D-n_{\mathrm{lang}}-i.
\end{equation}
For the corresponding pair $j=i$, this offset is
$-(D+n_{\mathrm{lang}})$ and changes with the prompt. The same image patch
therefore does not appear at one fixed rotary displacement across
instructions.

The anchored interface instead assigns
\begin{equation}
p_{i,\mathrm{anchored}}^{\mathrm{sp}}=D+i,
\qquad
\Delta_{ij}^{\mathrm{anchored}}=j-D-i.
\end{equation}
Every pair with the same spatial displacement $j-i$ now receives the same
relative rotation across image locations and instructions. In particular, the
corresponding pair $j=i$ always has offset $-D$. Anchoring thus provides a
prompt-invariant positional address from special token $i$ to visual position
$i$. A learned attention head can reuse the same relative-position pattern
along the full spatial diagonal.

This positional correspondence does not create the direct term present in the
vision-token interface. The mapping from a special-token output to a visual
input remains part of $J_{ij}^{A}$. Anchoring changes the positional structure
available to attention but does not add the same-position identity term
$\delta_{ij}I$. It therefore does not guarantee that attention or the
resulting gradient is concentrated on $v_i$. Content features, learned
projections, the softmax, and subsequent layers can strengthen, redistribute,
or suppress the route.

The derivation establishes index consistency rather than gradient alignment.
The same-region gradient measurements in the VLA-scale analysis test whether
the trained model actually uses this available route.

\subsection{Propagation through Depth and Encoder Parameters}

For an in-sequence implementation, let
$X^{(0)}=[V^\top,Q^\top,C^\top]^\top$ and consider pre-normalized residual
blocks
\begin{equation}
X^{(\ell+1)}=X^{(\ell)}+f_\ell(X^{(\ell)}),
\qquad
J_\ell:=\frac{\partial f_\ell(X^{(\ell)})}{\partial X^{(\ell)}}.
\end{equation}
Their end-to-end Jacobian can be written
\begin{equation}
\frac{\partial X^{(L)}}{\partial X^{(0)}}
=(I+J_{L-1})\cdots(I+J_0)=I+\mathcal{R},
\end{equation}
where $\mathcal{R}$ collects paths containing at least one contextual
transformation. Let $S_V$ and $S_Q$ select vision and special positions. The
two interface Jacobians are
\begin{equation}
B^{\mathrm{vis}}=I+S_V\mathcal{R}S_V^\top,
\qquad
B^{\mathrm{sp}}=S_Q\mathcal{R}S_V^\top,
\label{eq:app-deep-interface}
\end{equation}
because $S_QIS_V^\top=0$. Depth can add increasingly complex contextual routes
to either interface, but only the position-preserving vision stream contains
an explicit residual identity path between the same input and output
positions. A post-normalized architecture multiplies this route by
normalization Jacobians, and a fixed dimensionality-changing projection
replaces $I$ with its same-position Jacobian. Neither change creates an
identity route from a visual position to a distinct special-token position.

\paragraph{Encoder-parameter gradients.}

Let $J_j^E=\partial v_j/\partial\theta_v$ denote the encoder-parameter
Jacobian. The complete parameter update is
\begin{equation}
\frac{\partial\mathcal{L}}{\partial\theta_v}
=
\frac{\partial\mathcal{L}_{\mathrm{act}}}{\partial\theta_v}
+\lambda\sum_{j=1}^{N}(J_j^E)^\top g_j.
\label{eq:app-encoder-gradient}
\end{equation}
The interface changes encoder learning through the token-level signals $g_j$.
The decomposition does not order gradient magnitudes or downstream
performance. Contextual contributions may reinforce, redistribute, or cancel
the aligned component, and the encoder Jacobian further transforms every
token-wise contribution. These consequences are therefore evaluated
empirically rather than inferred from the structural expression alone.

\section{Limitations and Future Work}
\label{app:limitations}

Our study deliberately fixes the predictive target to a fixed-horizon residual
of momentum-encoded visual features. This isolates the effect of the prediction
interface, but it does not establish that one route is optimal for every
predictive objective. Targets differ in spatial granularity, temporal
abstraction, and relevance to action. Dense visual reconstruction, object
motion, geometric change, and semantic events may each benefit from different
ways of delivering supervision to the policy. Future work should therefore
study target and route jointly, including how prediction horizon and target
granularity determine which information remains useful for control.

We instantiate direct spatial coupling by predicting from every vision token.
This makes the same-position route explicit, but it is only one design within a
broader interface space. Sparse region-level or object-centric carriers could
retain explicit spatial addresses or residual links to selected visual tokens,
while allocating prediction capacity to the parts of the scene expected to
change. Such interfaces would test whether the observed transfer depends on
dense patch coverage or on the more general properties of stable correspondence
and direct coupling. Carrying these designs and their routing interventions
into jointly trained VLA policies would also extend the causal evidence beyond
the controlled encoder studied here.

\end{document}